\documentclass{article}

\usepackage{microtype}
\usepackage{graphicx}
\usepackage{subfigure}
\usepackage{booktabs} 

\usepackage{hyperref}

 \usepackage[accepted]{icml2025}

\usepackage{amsmath}
\usepackage{amssymb}
\usepackage{mathtools}
\usepackage{amsthm}

\usepackage[capitalize,noabbrev]{cleveref}
\theoremstyle{plain}

\theoremstyle{definition}

\theoremstyle{remark}

\usepackage[textsize=tiny]{todonotes}

\icmltitlerunning{Gaze Regularized Attention for VLMs}

\begin{document}

\twocolumn[
\icmltitle{
Gaze-Regularized VLMs for Ego-Centric 
Behavior Understanding}



\icmlsetsymbol{equal}{*}

\begin{icmlauthorlist}
\icmlauthor{Anupam Pani}{yyy}
\icmlauthor{Yanchao Yang}{yyy}
\end{icmlauthorlist}

\icmlaffiliation{yyy}{HKU Musketeers Foundation Institute of Data Science, The University of Hong Kong}

 \icmlcorrespondingauthor{Anupam Pani}{apani3@connect.hku.hk}

\icmlkeywords{Machine Learning, ICML}

\vskip 0.3in
]
\newcommand{\img}{\mathrm{I}}
\newcommand{\heat}{\mathrm{H}}
\newcommand{\gaze}{\mathrm{G}}

\newcommand{\tauo}{\tau_o}
\newcommand{\taua}{\tau_a}

\newcommand{\yc}[1]{{\textcolor{red}{Y: #1}}}
\newcommand{\ap}[1]{{\textcolor{blue}{A: #1}}}


\printAffiliationsAndNotice{}  

\begin{abstract}

Eye gaze, 
encompassing fixations and saccades, 
provides critical insights into human intentions 
and future actions. 
This study introduces a gaze-regularized framework that enhances Vision Language Models (VLMs) for egocentric behavior understanding. 
Unlike existing methods 
that rely solely on visual data and overlook gaze information, 
our approach directly incorporates gaze information 
into the VLM architecture during training. 
By generating gaze-based queries, 
the model dynamically focuses on gaze-highlighted regions, 
while a gaze-regularization mechanism ensures the alignment of model attention with human attention patterns. 
To better understand how gaze can be effectively integrated into VLMs, 
we conducted extensive experiments exploring various strategies for incorporating gaze data. 
These innovations enable the prediction of future events with detailed action descriptions. 
Experimental results demonstrate a nearly 13 $\%$ improvement in semantic scores compared to baseline models not leveraging gaze data, highlighting the effectiveness of our approach. 
This work establishes a foundation for leveraging the human gaze in VLMs, significantly boosting their predictive capabilities in applications requiring accurate and robust future event prediction.
\end{abstract}
\section{Introduction}
\label{sec:intro}

\begin{figure*}[!t]
    \centering
    \includegraphics[width=0.9\linewidth]{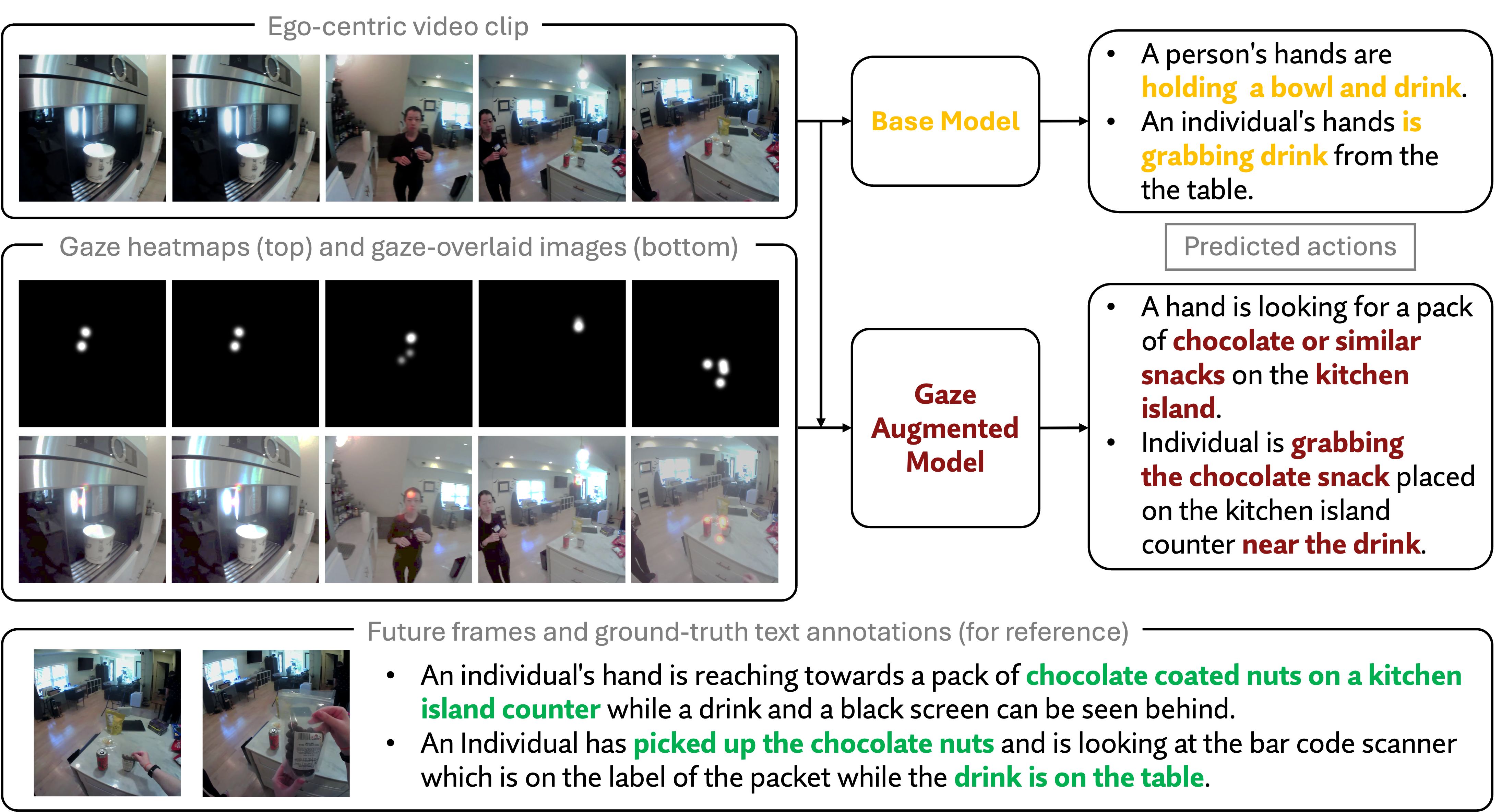}
    \vspace{-3mm}
    \caption{Illustration of future event prediction. 
    The input is a sequence of image frames, 
    and the output is an action-related text prediction. 
    Unlike the gaze-regularized model, the base model misidentifies the object to be picked up (bowl). 
    Predicted future annotations for both models are shown on the right, with ground-truth annotations and immediate future frames displayed below for reference.}
    \label{fig:problem-statement}
\end{figure*}


VLMs are foundation models 
that combine computer vision and natural language processing 
to understand and generate both visual and textual information. 
Examples of such models 
include ViLBERT, LXMERT, and CLIP \citep{lu2019vilbertpretrainingtaskagnosticvisiolinguistic,tan2019lxmertlearningcrossmodalityencoder,radford2021learningtransferablevisualmodels}. 
These models can generate textual descriptions from images and vice-versa. 
When adapted for predicting future events and generating descriptive captions, VLMs hold significant promise for advancing human-machine interaction across various applications
like assistive robots \citep{li2024visionlanguagefoundationmodelseffective}, 
accessibility for visually impaired individuals \citep{zhao2024vialmsurveybenchmarkvisually}, 
and autonomous driving \citep{zhou2024visionlanguagemodelsautonomous}, 
contributing to safer and more inclusive environments.

To fully realize their potential,
we propose that VLMs should be equipped with the ability to understand egocentric behavior and generate fine-grained predictions,
providing more actionable information to enhance human-machine interaction.
While coarse predictions might identify general events like ``brewing coffee,'' 
fine-grained predictions specify detailed steps such as ``reaching for the coffee capsule in the top-right cabinet, then filling the tank with the cup on the rack.'' 
These fine-grained predictions
provide useful insights that allow machines and humans 
to interact more effectively with each other \citep{goyal2021annotatingmodelingfinegrainedfactuality}. 

Achieving accurate fine-grained predictions and detailed descriptions requires understanding the human agent's short-term goals, which, we propose, can be inferred from eye gaze patterns. Eye gaze provides insights into the objects a person is focusing on or intends to interact with, making it a critical component for improving future event prediction
\citep{Frischen2007,Tipper2010}. 
Moreover, humans often scan their surroundings to achieve a better understanding of their environment, using eye gaze to gather detailed information about spatial relationships, object properties, and potential interactions with the objects around them.

Thus, integrating eye gaze into VLMs enables fine-grained future activity prediction and current activity understanding by providing deeper insights into a person’s focus and intentions. This study introduces a gaze-regularized framework that leverages gaze patterns to understand egocentric behavior.
By generating gaze-based queries to enhance scene features,
 our framework dynamically directs the model’s attention to regions of interest. Specifically, it employs an attention mechanism with a Vision Transformer (ViT) \citep{dosovitskiy2021imageworth16x16words}, and includes gaze-based queries while RGB image features serve as keys and values. To ensure alignment between human gaze and model attention, we introduce a gaze-regularization mechanism that balances model-based attention with human gaze patterns.
Our experiments demonstrate that incorporating gaze information significantly boosts performance, achieving nearly a 13 $\%$ improvement in semantic scores for future activity predictions compared to baseline models and a 2$\%$ improvement in understanding and describing the current activity.
Further testing with detailed annotations shows an additional 12 $\%$ performance increase in gaze-augmented models for future event prediction, while baseline models remain unchanged, highlighting the importance of fine-grained annotations for maximizing the framework's potential and the value of incorporating gaze into the system.
In summary, our contributions are 
1) A set of gaze-regularized VLMs 
for accurate egocentric behavior understanding; 
2) A novel gaze-regularized attention mechanism 
that guarantees effective training of the gaze-augmented VLMs;
3) An efficient gaze sanity check to ensure the relevance of the gaze information; 
and 4) An extensive study with experiments that demonstrates the effectiveness of the proposed framework and validates the significance of gaze information and its inclusion in VLMs.

\section{Related Work}

\textbf{Attention and Gaze-Augmented Models}:
Attention-based mechanisms have been widely used to identify salient features in vision tasks, improving action recognition and human-computer interaction \citep{8082802, 9340893, 8793804, 9473584}. Gaze information has been employed to guide attention maps for activity recognition, either as explicit supervision \citep{9022139, Min, awale2022human} or as an auxiliary signal. Prior works have leveraged gaze for next-active object prediction \citep{thakur2023enhancingactiveobjectbasedegocentric} and task-specific information extraction in real-world interactions \citep{Hayhoe2003-gy}. In our work, we incorporate gaze using our proposed gaze-regularized attention framework to enhance egocentric behavior understanding.

\textbf{Action Prediction and Forecasting}:
 Prior work has explored LSTM-based models and modality attention mechanisms to anticipate actions from egocentric videos \citep{furnari2019expect, gao2017red}. Hierarchical modeling approaches \citep{ashutosh2023hiervllearninghierarchicalvideolanguage} have been introduced to capture both immediate actions and high-level goals, while goal-driven methods \citep{zhao2024antgptlargelanguagemodels, mascaro2024intentionconditioned} decompose predictions into low-level actions conditioned on high-level intentions.Other studies emphasize object-hand interactions \citep{cho2024shorttermobjectinteractionanticipation} or model spatiotemporal dependencies through transformer-based architectures \citep{roy2024interactionregionvisualtransformer, lai2024legolearningegocentricaction}. Our work extends these approaches by incorporating gaze and providing fine-grained text annotations.
More information about prior work can be found in Sec.~\ref{sec:more-related-work} in the appendix.
\section{Method}
\label{sec:method}



Our goal is to develop a VLM 
that can leverage gaze information 
to understand egocentric behavior 
through two key tasks: 
predicting future activities and 
understanding the current activity. 
By aligning model attention with gaze patterns, 
the framework is capable of 
inferring human intention critical 
for future event prediction 
and context-aware descriptions of ongoing actions. 
In the following, 
we define the problem, 
outline the training data, 
and present the proposed gaze-regularized attention mechanism to
enhance the model’s performance across both tasks.

\paragraph{Problem setup} 
\label{intial-setup}
Let $\phi_{gaze}$ represent the gaze-regularized VLM, 
designed to model the likelihood of fine-grained text descriptions ($\ell_i$) by incorporating both visual and gaze information. 
Given a sequence of egocentric video frames $\{\img_i\}_{i=1}^{\tauo}$ over the past $\tauo$ seconds (where $\tauo$ is observation time), the model performs the following two tasks:

1) \textbf{Future Activity Prediction}: 
Generating text descriptions $\{\ell_i\}_{i=\tauo+1}^{\tauo+\taua}$ corresponding to future frames (what will happen) in the imminent $\taua$ seconds (anticipation time). The formulation of this task is:
\begin{equation}
\begin{split}
    \phi_{gaze}(\ell , \{\img\},\{\heat\},\{\gaze\}) = 
    \prod_{i=\tau_o+1}^{\tau_o + \tau_a} 
    p(\ell_i \mid \ell_{ < i}, \img_{\leq \tau_o}, \\
    \heat_{\leq \tau_o}, \gaze_{\leq \tau_o})\,,
\end{split}
\end{equation}
2) \textbf{Current Activity Understanding}: 
Generating fine-grained text descriptions $\{\ell_i\}$ for the observed frames (what is happening). This task can be formulated as:
\begin{equation}
 \begin{split}
    \phi_{gaze}(\ell,\{\img\},\{\heat\},\{\gaze\}) = 
    \prod_{i=1}^{\tau_o} 
    p(\ell_i \mid  \img_{\tau_o}, 
    \heat_{\tau_o}, \gaze_{\tau_o})\,,
 \end{split}
\end{equation}
where $\{\heat\}$ represents the gaze heatmaps, 
and $\{\gaze\}$ refers to gaze-overlaid or pseudo-gaze-overlaid images (with gaze prediction), created by blending gaze heatmaps with the corresponding image frames.

Human brains predict future events 
over hierarchical timescales, 
as noted by \citet{lee2021anticipation}. 
These timescales range from fine-grained changes, such as words or images over 1–4 seconds, 
to coarse-grained evolutions, like plotlines spanning up to 15 seconds. 
We focus on fine-grained activity and event anticipation to enhance human-machine interaction, setting the anticipation time ($\taua$) to 2 seconds by default.
Additionally, research by \citet{Seidel2014} shows that attention control mechanisms are usually activated in response to stimuli or duress and operate up to 2 seconds before an event occurs. 
Based on this, we set the observation time ($\tauo$) for the future prediction task to more than 2 seconds, capturing meaningful behavioral patterns. 
For the activity understanding task, 
an observation time of 1 second is used to focus on the immediate context.

\subsection{Dataset Construction}

We use the Ego4D dataset \citep{grauman2022ego4d}, 
which includes video clips with eye-gaze data. 
Gaze points are converted 
into images to highlight important visual regions. 
Videos are downsampled to one frame per second to reduce computational load.
To support both current activity understanding and future event prediction tasks, 
we create detailed textual descriptions using GPT-4V \citep{openai2023}. 
This involves generating initial frame-by-frame descriptions and 
refining them by iteratively improving the prompt based on manual feedback. 
The final text annotations 
provide contextually coherent descriptions 
for each frame. Specifically, 
for every image frame $\{\img_i\}$, 
we obtain a corresponding text annotation $\{l_i\}$.
During model training and testing, 
the input consists of a sequence of observed images $\{\img_i\}_{i=1}^{\tauo}$, while the textual annotations $\{l_i\}_{i=\tauo+1}^{\tauo+\taua}$ corresponding to the image frames $\{\img_i\}_{i=\tauo+1}^{\tauo+\taua}$ are used as ground truth for future activity prediction. 
For each observed image $\{\img_i\}$, 
a text annotation $\{l_i\}$ is also provided to support the activity understanding task. 
Figure~\ref{fig:prompt-design1} and Figure~\ref{fig:prompt-design2} in the Appendix provide a diagram illustrating this process,
along with an example of the resulting annotations (Figure~\ref{fig:prompt-example}). Please refer to Sec.~\ref{sec:app_ds_construction} for more details about dataset construction.



\subsection{Base Model}
Leveraging Flamingo's \citep{alayrac2022flamingo} capability 
to model cross-modal interactions between 
images and language, 
we adapt the open-source Flamingo model \citep{awadalla2023openflamingo} 
as our base model. 
We train the perceiver resampler in the vision block and the cross-attention layers in the language block, 
keeping other components frozen. 
The input image sequence 
$\{\img_i\}_{i=1}^{\tauo}$ 
is processed by a pre-trained ViT to extract visual features, 
which are then fed into a perceiver resampler, 
learning a fixed-size representation of the visual data and leveraging time embeddings to capture temporal relationships.
The output from the perceiver resampler connects to the language module to generate text annotations. 
This base model, trained without gaze information,
serves as a benchmark for evaluating our gaze-regularized models for future activity prediction and activity understanding tasks.
Let $\phi_{base}$ represent the base model which does not utilize gaze, 
then the \textbf{future activity prediction} task can be formulated as follows: 
\begin{equation} 
\phi_{base}(\ell, \{\img\}) = 
\prod_{i=\tau_o+1}^{\tau_o + \tau_a} 
p(\ell_i \mid \ell_{ < i}, \img_{i\leq\tau_o})\,. 
\end{equation}
And the \textbf{activity understanding} task can be formulated as:
\begin{equation}
 \begin{split}
    \phi_{base}(\ell,\{\img\}) = 
    \prod_{i=1}^{\tau_o} 
    p(\ell_i \mid  \img_{\tau_o})\,,
 \end{split}
\end{equation}
Furthermore, 
let $P(\ell)$ be the actual probability distribution of the ground-truth text, 
and $\phi_{base}(\ell,\{\img\})$ 
the model predicted distribution. 
The cross-entropy loss to be minimized for training can be written as: 
\begin{equation} 
\mathcal{L}_{\mathrm{CE}}(\{\img\},\ell) = 
-\sum P(\ell)\log(\phi_{base}(\ell , \{\img\}))\,. \end{equation}
This setup provides a clear baseline, enabling us to compare and assess the impact of incorporating gaze information in the subsequent gaze-regularized models.



\subsection{Gaze-Regularized Model}
\begin{figure*}[!t]
  \centering
    \includegraphics[width=1\textwidth]{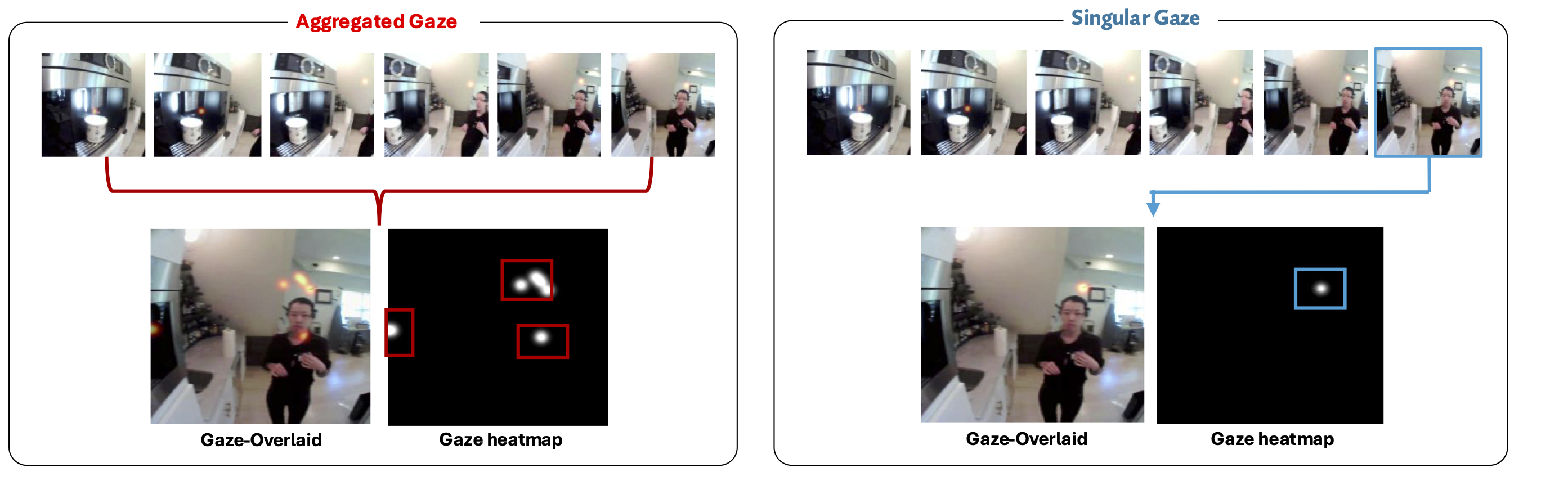}
        \vspace{-6mm}
  \caption{
  \textbf{Heatmap creation.} 
  Illustration of gaze data collection 
  for generating gray-scale heatmaps and gaze-overlaid images. 
  On the left, the aggregated gaze model incorporates multiple gaze points collected over the interval $[t-\delta, t]$ to generate the heatmap. 
  On the right, the singular gaze model uses a single gaze point collected at time $t$. Both utilize Gaussian smoothing to generate the heatmap.
  }
  \label{fig:gaze-aggregation}
\end{figure*}

Human gaze patterns reveal critical insights into attention and focus, providing clues about intentions, thoughts, and potential actions \citep{Tipper2010,Frischen2007}. 
Leveraging this natural attention mechanism, we propose a gaze-regularized attention block to enhance visual features from the pre-trained ViT before passing them to the Perceiver Resampler. 
The model input includes a sequence of RGB images $\{\img_i\}_{i=1}^{\tauo}$ whereas during training corresponding binary heatmaps $\{\heat_i\}_{i=1}^{\tauo}$ highlight gaze regions. 
These heatmaps are generated from textual eye gaze data and are blacked out except for the gaze points, which are smoothed using a Gaussian filter to create a heatmap. Moreover, 
these heatmaps generate a gaze distribution, 
indicating how gaze is distributed across image patches. 
Unlike the base model, the gaze-regularized model integrates both eye gaze data and RGB images to improve performance. Examples of binary heatmaps and gaze-overlaid images derived from the eye gaze data are shown in Figure~\ref{fig:gaze-aggregation}. Additional details about the proposed gaze-regularized attention model and it's components are provided in Sec.~\ref{sec:overview-appendix} of the Appendix.

\begin{figure*}[!t]
  \centering
    \includegraphics[width=1\textwidth]{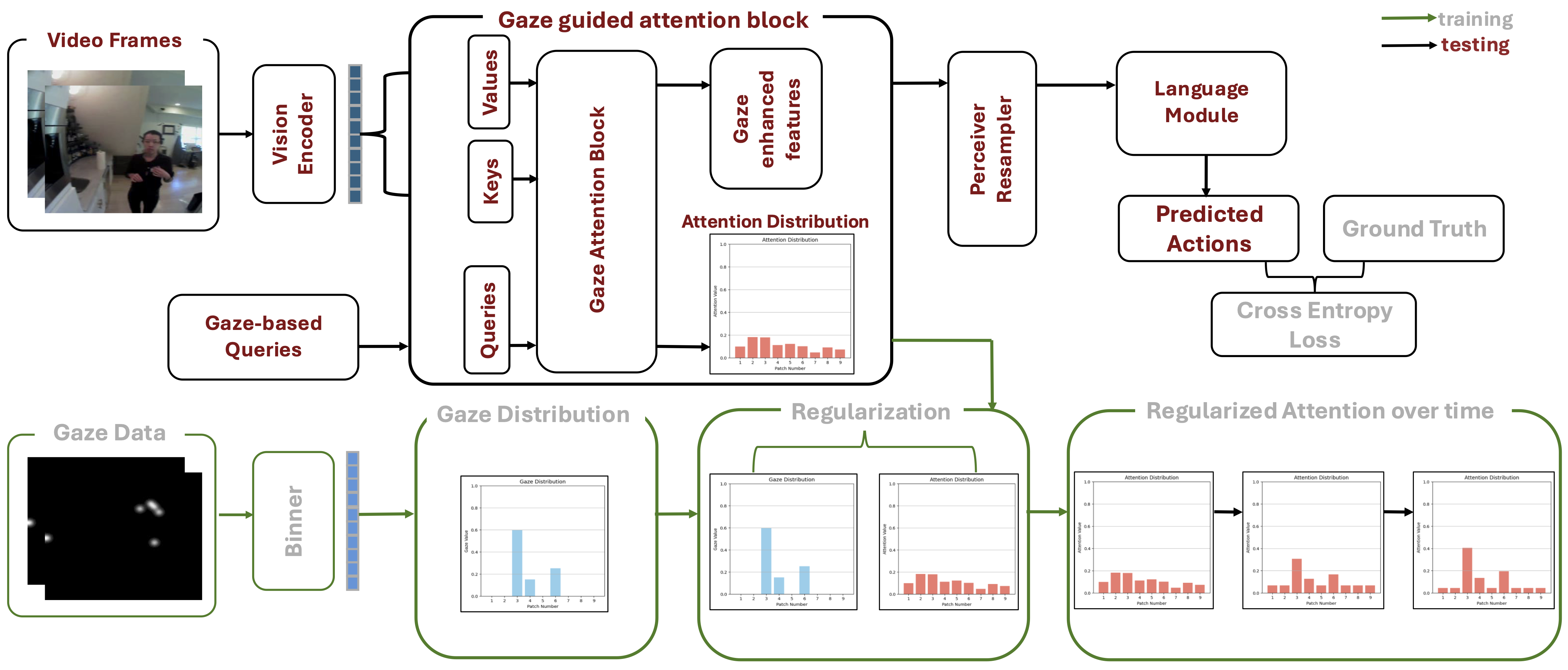}
    \vspace{-6mm}
  \caption{\textbf{Overview of the Architecture} 
  A ViT encoder extracts image features, 
  which are enhanced using gaze-based queries 
  (obtained from gaze/psuedo gaze-overlaid images) 
  in a gaze-regularized attention block. 
  A Perceiver Resampler generates a fixed-size representation for the language module to predict text annotations. 
  The gaze regularizer aligns model attention with human gaze patterns by minimizing the Kullback–Leibler divergence between the model attention distribution and gaze distribution during training.}
  \label{fig:vision-gaze-model}
\end{figure*}

\subsubsection{Gaze distribution computation}
Each binary heatmap image is first divided into patches. 
Since the gaze is usually concentrated on specific regions, 
not every patch will contain gaze information. 
Pixels occupied by gaze are indicated by non-zero pixel values, 
while pixels without gaze are blacked out, having values of zero.
For a binary heatmap image denoted as $\heat_t$, 
let $N$ represent the total number of patches. 
The proportion of gaze occupied within a patch $N_{i,j}$, where $(i,j)$ represents the patch's position in the grid, can be calculated as follows:
\begin{equation}
N_{i,j} = \frac{\sum_{y=\frac{j(h)}{n_v}}^{\frac{(j+1)(h)}{n_v}} \sum_{x=\frac{i(w)}{n_h}}^{\frac{(i+1)(w)}{n_h}} p_{xy} } {\sum_{y=0}^{h} \sum_{x=0}^{w} p_{xy}},
\end{equation}
with $i \in [0, n_h - 1]$ and $j \in [0, n_v - 1]$.
Here, $h$ and $w$ represent the height and width of the image, 
$n_h$ and $n_v$ denote the number of horizontal and vertical patches, 
respectively, and $p_{xy}$ is a binary variable representing the pixel value at position $(x,y)$, 
which can be either 0 or 1. 
The denominator corresponds to the total value of all pixels in the image.
For each binary heatmap image $\heat_t$, 
we derive a gaze distribution $\hat{H_t}$, 
which is a vector of shape $(1, n_h \times n_v)$. 
The number of patches, $N$, in the gaze distribution, is equal to the number of tokens produced per image by the ViT. 
This gaze distribution plays a crucial role in the gaze regularizer used in the attention block, 
acting as the target distribution that we want the attention distribution to mimic, which will be discussed in the following section.

\subsubsection{Gaze-regularized attention block}
To obtain features for the gaze-regularized attention block, we use a pre-trained ViT to process both RGB images and gaze-based queries. These gaze-based queries can be obtained through two approaches, depending on the availability of gaze-overlaid images during deployment:

1) \textbf{Gaze-overlaid images present during test time}: Features are directly extracted from gaze-overlaid images using the ViT (refer to Sec.~\ref{sec:vit-details} in the Appendix for more details). These features serve as queries ($Q$) in the gaze-regularized attention block. 

2) \textbf{Gaze-overlaid images absent during test time}: 
During training, 
we use gaze heatmaps as supervision 
to train a separate module that predicts heatmaps. 
During test, 
the predicted heatmaps 
are dynamically combined with RGB images to create pseudo-gaze-overlaid images, allowing the model to produce gaze-based queries without relying on the availability of gaze data at test time. 
A cosine-similarity loss is applied during training to ensure the generated pseudo-gaze-overlaid images closely match the ground-truth gaze-overlaid images.
More details about this process can be found in Sec.~\ref{sec:pseudo-gaze-overlaid} and Figure.~\ref{fig:gaze-overlaid-w-supervision} in the Appendix.

Simultaneously, 
the image features from the RGB images are used as keys ($K$) and values ($V$). 
Attention is then calculated based on the following equation:
\begin{equation} 
\text{Attention}(Q, K, V) = \text{softmax}\left(\frac{QK^T}{\sqrt{d_k}}\right)V = A V \,,
\end{equation}
where $A$ represents the attention weights. 
Simply using gaze features 
as queries does not guarantee that the attention will be concentrated on gaze-allocated regions. 
The attention scores will be distributed across the image. 
We aim to guide the model to prioritize these regions, 
directing more focus (attention) 
towards areas where the gaze is present. 
To this end, 
we introduce a gaze regularizer designed to enhance the model’s attention on gaze-allocated regions. 
The gaze regularizer is implemented using Kullback-Leibler (KL) divergence, 
which is applied to the training of the gaze-regularized model. 
The aim of the regularizer is to ensure that the distribution of the attention weights is more closely aligned with the gaze distribution obtained from the binary heatmap image. 
We denote the attention weight distribution as $A$, 
while the gaze distribution is represented by $H$. 
The KL divergence for the two distributions can be obtained using the following equation:
\begin{equation} 
D_{KL} (A \Vert H) = \sum_{i}{A}_i \log{\frac{{A}_i}{H_i}}\,.
\end{equation}
The binary heatmap images 
are not used as queries because doing so could lead to important image information being overlooked. 
For instance, 
the heatmap images for two totally different images could be the same if one looks at the center of the image.
To avoid this, 
we use gaze-based queries, 
which mimic the gaze information while also retaining the unique visual content of each image. 
It is crucial to preserve the visual content while aligning the attention distribution with the gaze distribution to ensure the model properly integrates both sources of information.
The overall objective function we minimize to train the gaze-regularized model is as follows:
\begin{equation} 
\mathcal{L}_{\text{total}} 
= \mathcal{L}_{\text{CE}} + \lambda * {D_{KL}} (A \Vert H)\,,
\end{equation}
where $\lambda$ is the coefficient of the regularization term in the proposed objective function, 
and the cross-entropy loss is similar to the loss from the base model. 
By utilizing this gaze-regularized attention mechanism within the model, 
we can 
orient the model's attention 
to regions highlighted by gaze, 
and
improve the performance of the proposed VLM.

\subsubsection{Singular and Aggregated gaze regularized models}

In the eye-tracking literature, 
eye movements are classified into fixations and saccades. Fixations, lasting around 200 ms \citep{rayner200935th}, allow the eyes to pause and gather detailed visual information from a specific region. 
Though minor involuntary movements occur, the focus remains on a fixed area. 
In contrast, saccades are rapid eye movements that shift gaze between points of interest, introducing significant visual changes. As saccades enable efficient visual search, two consecutive saccades can cover large distances.
Given the nature of saccades and fixations, as well as their average durations, detailed or relevant information is typically collected within a temporal window of $\delta=200$ milliseconds. 
Correspondingly, 
we provide both a singular gaze-regularized model and an aggregated gaze-regularized model. 
For each image frame $\img_t$, 
we generate a heatmap image $\heat_t$ using gaze coordinates available at time $t$. 
Gaze data is sampled at 30 frames per second and stored in the format $g_t=(\text{timestamp}, x, y)$, 
where $(x, y)$ is the coordinate of the gaze point. 
For the singular gaze-regularized model, 
the heatmap at time $t$, $\heat_t$, is:
\begin{equation} 
\heat_t = f(g_t)\,. 
\end{equation}
Here, the function $f$ represents a mechanism to construct the heatmap, 
such as a Gaussian smoothing. 
In the singular gaze-regularized model, 
we use a single gaze point captured at the timestamp nearest to when the RGB image is obtained. 
However, a singular gaze point can be noisy due to measurement errors or micro-saccades \citep{Rolfs2009uv,Ratliff_1950,Collewijn_2008}. 
An aggregated gaze pattern can account for such variations. 
To minimize the impact of noise and ensure a sufficient time frame for collecting detailed information, 
we propose aggregating points within a fixed time interval $\delta$ around the image frame $\img_t$. 
The corresponding heatmap $\heat_t$ at time $t$ is then:
\begin{equation} 
\heat_t = \sum f({g_i}), \quad \forall i \in [t-\delta, t]\,, 
\end{equation}
which is normalized by the length of the interval, 
and ${g_i}$ for all $i \in [t - \delta, t]$ represents the set of gaze points lying within the vicinity of timestamp $t$. 
These gaze points are used to obtain the binary heatmap image $\heat_t$.
However, aggregation can sometimes include gaze points that may be occluded in the final frame. 
To address this, we implement an occlusion check using forward and backward optical flow consistency.
The purpose of the occlusion check is to ensure that if a significant occlusion occurs, 
the corresponding gaze points are excluded from the heatmap creation process. 
More information about the occlusion check can be found in Sec.~\ref{sec:occlusion-check-correction}. 
An example of using the occlusion check is shown in Figure~\ref{fig:occlusion-check}.
Next, we demonstrate the effectiveness of the proposed gaze-regularization mechanisms for egocentric behavior understanding.

\section{Experiments}
\label{sec:exp}

This section evaluates base models against gaze-regularized models using semantic and ROUGE-L scores, 
highlighting the impact of gaze-regularized attention. 
We analyze different variations 
to identify the best configuration and demonstrate that current VLMs struggle with gaze utilization and fine-grained future activity prediction. 
Additional experiment details are in Sec.~\ref{sec:other-exps} in the Appendix.

\begin{figure*}[t]
  \centering
    \includegraphics[width=0.95\textwidth]{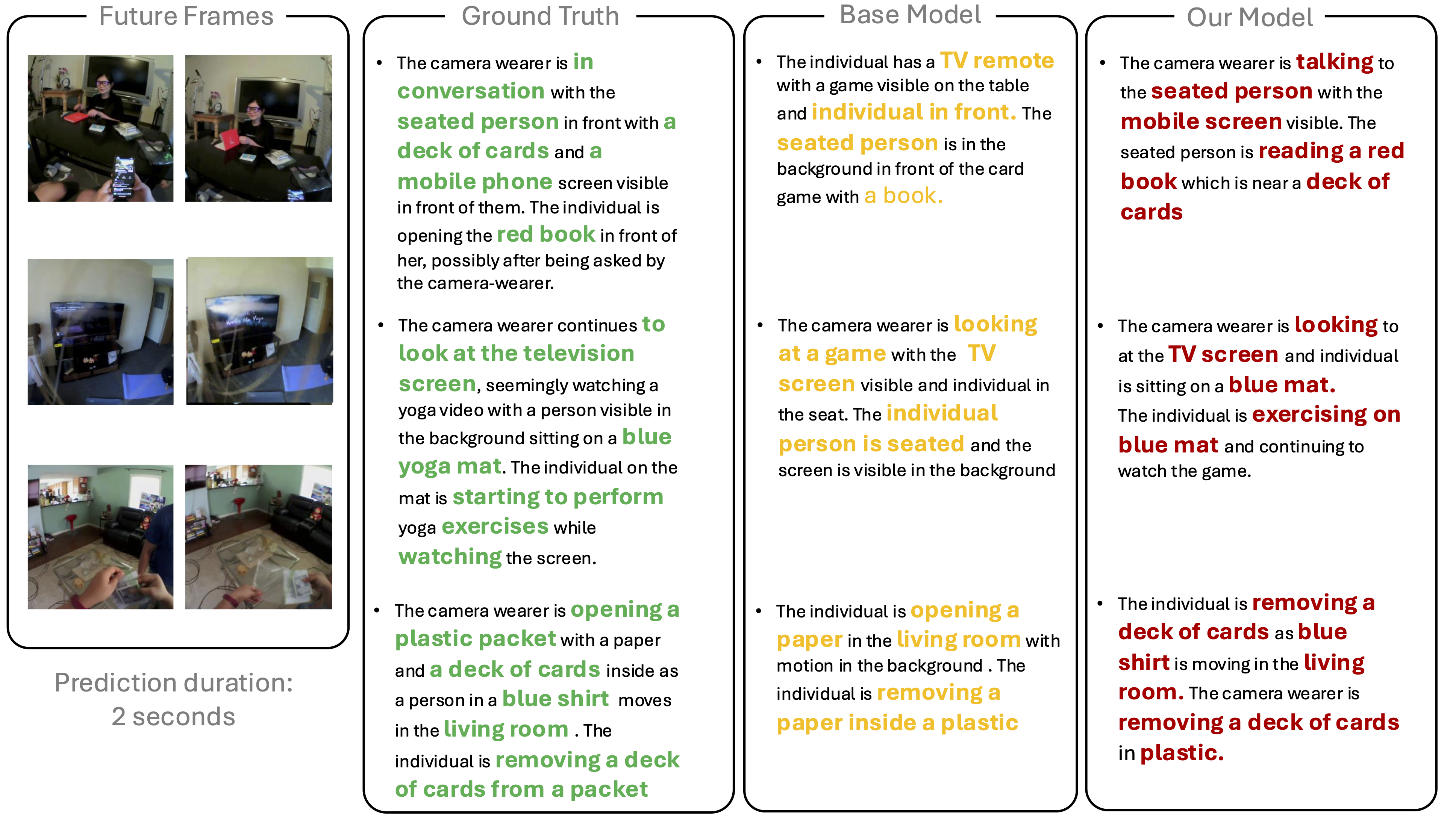}
    \vspace{-4mm}
  \caption{Future event prediction results for the base model (without gaze) and our gaze-regularized model are presented for an observation horizon ($\tauo$) of 5 seconds. Past frames are omitted, but ground-truth annotations and future frames with a prediction duration ($\taua)$ of 2 seconds are provided as references. Keywords from each set of annotations are highlighted for easier reading}
  \vspace{-2mm}
  \label{fig:pred-results-exp}
\end{figure*}

\subsection{Comparison between Base Model and Gaze-Regularized Models}
The objective of this experiment is to quantify the impact of eye gaze data, especially on future egocentric activity prediction 
by comparing model performance using measurable metrics. 
The base model receives only RGB image frames as input, 
while the gaze-regularized models incorporate multiple variations of gaze information. 
All gaze-regularized models 
share the same architecture but differ in the way gaze data is utilized.
In the Singular Gaze-regularized Model, 
for each image frame $\img_t$ at time $t$, the corresponding heatmap $\heat_t$ is generated using a single gaze point $g_t$, obtained from the timestamp closest to $t$. 
In the Aggregated Gaze-Regularized Model, for each image frame $\img_t$ at time $t$, the corresponding heatmap $\heat_t$ is generated using aggregated gaze points ${g_k}$, where $k \in [t-\delta, t]$ and $\delta$ is set to 200 milliseconds.

All models are trained under identical conditions. As shown in Table~\ref{tab:result1}, 
the gaze-regularized models outperform the base model across all evaluation metrics. 
Using aggregated gaze points over a 200-millisecond interval shows a noticeable improvement in performance compared to using a single gaze point. 
This can be attributed 
to the larger focus area and the temporal gaze pattern enabled by the aggregation, 
allowing the model to capture a broader context for inferring the intent 
and better facilitate the anticipation of future activities. 
Please refer to Figure~\ref{fig:pred-results-exp} for qualitative results.

\begin{table}[t]
\vspace{-3mm}
\caption{Evaluation of base and gaze-regularized models for future prediction with varying annotation granularity (i.e., GPT-4V and ShareCaptioner).}
\vspace{-1mm}
\label{tab:result1}
\begin{center}
\begin{scriptsize} 
\begin{sc}
\begin{tabular}{lccc}
\toprule
Model               & Annotation  & Semantic  & Rouge-L  \\
                    & Source & score ($\uparrow$) & F-score \\
\midrule
Base (no gaze)      & GPT-4V            & 0.6525  & 0.4318          \\
Singular Gaze       & GPT-4V            & 0.7316  & 0.5054          \\
\textbf{Aggregated-Gaze} & GPT-4V            & \textbf{0.7826} & \textbf{0.5405} \\
\midrule
Base (no gaze)      & ShareCaptioner    & 0.6437  & 0.5646          \\
Singular Gaze       & ShareCaptioner    & 0.8212  & 0.6905          \\
\textbf{Aggregated-Gaze} & ShareCaptioner    & \textbf{0.9125} & \textbf{0.7666} \\
\bottomrule
\end{tabular}
\end{sc}
\end{scriptsize} 
\end{center}
\vspace{-4mm}
\end{table}

\subsection{Effect of the Gaze-Regularizer}

We evaluate the impact of the gaze regularizer in the aggregated gaze-regularized model by testing different values of $\lambda$. When $\lambda = 0$, the regularizer is inactive, while larger values increase its influence. As shown in Table~\ref{tab:result2}, performance without regularization matches the base model, while optimal results occur at moderate $\lambda$ values, with slight declines at higher levels. 
The diminished performance without regularization underscores the significance of aligning attention with the gaze distribution through the gaze regularizer when the gaze attention block is included.


\begin{table}[t]
\vspace{-2mm}
\caption{Effect of regularization on aggregated gaze-regularized models.}
\label{tab:result2}
\begin{center}
\begin{scriptsize}
\begin{sc}
\begin{tabular}{lcc}
\toprule
Regularization  & Semantic  & Rouge-L  \\
Coefficient  & Score($\uparrow$)   & F-score($\uparrow$)    \\
\midrule
0        & 0.6317          & 0.3738          \\
\textbf{100} & \textbf{0.7826} & \textbf{0.5405} \\
1000     & 0.7798          & 0.5330          \\
\bottomrule
\end{tabular}
\end{sc}
\end{scriptsize}
\end{center}
\vspace{-2mm}
\end{table}

\subsection{Impact of the Annotation Quality on Model Performance}
 We designed a prompt-based method using GPT-4V to generate text annotations for image frames and supplemented them with ShareCaptioner \citep{chen2024sharegpt4video}, which produces more fine-grained captions capturing frame-to-frame changes. 
 Comparing both annotation types, we found that while the base model performed similarly with both, the gaze-regularized model improved by nearly 12$\%$ with the more detailed annotations (Table~\ref{tab:result1}). This suggests that gaze information becomes even more valuable when annotations are highly descriptive, as it helps the model focus on relevant regions for better action prediction

\subsection{Evaluating Gaze-based Query Generation Methods}
In this experiment, we evaluate four approaches to generate gaze-based queries in the gaze-regularized attention block. These approaches address the challenge that gaze-overlaid images may not always be available during testing. 

\textbf{Gaze-Overlay-Dependent Model}:
During training, this model uses RGB scene images alongside gaze-based queries derived from gaze-overlaid images. However, during testing, only RGB scene images are available, requiring the model to generate gaze-based queries directly from these images.

\textbf{RGB as Gaze-Queries}:
This model operates exclusively on RGB scene images during both training and testing. Gaze-based queries are generated directly from RGB images, without any reliance on gaze-overlaid images. Unlike the base model, this approach incorporates the additional gaze-regularized attention block before the perceiver resampler.

\textbf{Gaze-Overlay-Supported Model}:
This approach assumes the availability of gaze-overlaid images during both training and testing. RGB scene images and gaze-based queries derived from gaze-overlaid images are used in all phases, allowing the model to fully leverage gaze information.

\textbf{Pseudo-Gaze-Overlay-Supported Model}:
During training, this model uses gaze-overlaid images as supervision to train a module that generates heatmaps. These heatmaps are dynamically combined with RGB scene images to create pseudo-gaze-overlaid images for generating gaze-based queries. During testing, the model relies only on RGB scene images to dynamically generate pseudo-gaze-overlaid images which are then used as gaze-based queries, overcoming the need for precomputed gaze-overlaid images.

From Table.~\ref{tab:gaze-overlaid-check-test-time}, we observe that while the Gaze-Overlay-Supported Model achieves the best performance, it relies on gaze-overlaid images during training and test time. In contrast, the Pseudo-Gaze-Overlay-Supported Model offers a practical alternative, recreating the benefits of gaze-overlaid images without requiring them during test time and only utilizing gaze-overlaid images during training.

\begin{table}[t]
\caption{Comparison of various gaze-based querying methods.}
\vspace{-2mm}
\label{tab:gaze-overlaid-check-test-time}
\vskip 0.15in
\begin{center}
\begin{scriptsize}
\begin{sc}
\begin{tabular}{lcc}
\toprule
Gaze-based                     & Semantic       & Rouge-L \\
    Queries                                      & Score ($\uparrow$)       & F-score($\uparrow$)    \\
\midrule
Gaze-Overlay-Dependent   & 0.7368             & 0.5034          \\
RGB-Only  & 0.7505             & 0.5173          \\
\textbf{Gaze-Overlay-Supported } & \textbf{0.7826}    & \textbf{0.5405} \\
 Pseudo-Gaze-Overlay-Supported & 0.7704    & 0.5213 \\
\bottomrule
\end{tabular}
\end{sc}
\end{scriptsize}
\end{center}
\vskip -0.15in
\end{table}

\subsection{Performance Comparison Between Models for Activity Understanding}


We examine whether a static scene, independent of sequential context, is sufficient for the gaze-regularized models to understand an ongoing activity, noting that VLMs are often used in static-scene setting.
While a single image lacks temporal context, it contains visual cues and object affordances that hint at probable actions.
Unlike future prediction, which benefits from evolving gaze patterns, static scene understanding relies mainly on visible objects. If gaze is beneficial, it should highlight key regions that aid activity understanding.
To test this, we compare the base and gaze-regularized models in generating fine-grained descriptions from single images, isolating the impact of gaze without temporal cues. 
Results show that the gaze-regularized model achieves a 2 $\%$ improvement over the base model, indicating that gaze can enhance scene understanding even in single-image settings. However, its full potential is realized in tasks where evolving gaze patterns provide richer contextual cues.

\begin{table}[t]
\vspace{-3mm}
\caption{Comparison between base model and gaze-regularized models (with different regularization coefficients) for activity understanding tasks.}
\vspace{-2mm}
\label{tab:captioning-task}
\vskip 0.15in
\begin{center}
\begin{scriptsize}
\begin{sc}
\begin{tabular}{lcc}
\toprule
Model                    & Semantic       & Rouge-L \\
                                          & Score ($\uparrow$)       & F-score($\uparrow$)    \\
\midrule
Base  & 0.6897             & 0.4832          \\
Gaze-aggregated ($\lambda$ = 100)  & 0.6843             & 0.4816          \\
Gaze-aggregated ($\lambda$ = 1000)  & 0.6621            & 0.4793          \\
\textbf{Gaze-aggregated ($\lambda$ = 10)}  & \textbf{0.7021}             & \textbf{0.5001 }         \\
\bottomrule
\end{tabular}
\end{sc}
\end{scriptsize}
\end{center}
\vskip -0.15in
\end{table}

\subsection{Evaluating the Predictive Limitations of Pre-trained VLMs with Gaze-Augmented Data}
We conducted an additional experiment to evaluate the predictive capabilities of current VLMs (InternVL \citep{chen2024internvl}, Open-LLaVA-NeXT \citep{chen2024open} and OpenFlamingo \citep{awadalla2023openflamingo}) 
and the effect of incorporating gaze data 
without the proposed gaze regularization. 
Using pre-trained VLMs in a zero-shot prediction setting, we assessed their ability to generate future annotations for a sequence of images. This experiment highlights two key points:
(1) existing VLMs, designed for static image-text alignment and scene understanding, struggle with future event prediction, and 
(2) including gaze data as input does not significantly improve their performance, as these models lack the mechanisms to effectively leverage such information. 
This evaluation is not a direct comparison with our gaze-regularized models but serves to underscore the advantages of explicitly training models for gaze-driven predictions and how we can effectively utilize gaze as an additional signal in the vision language models.


\begin{table}[t]
  \caption{Comparison of semantic scores for other VLM Models with different gaze inputs on our test set.}
  \label{tab:semantic_scores}
\begin{center}
\begin{small}
\begin{scriptsize}
\begin{sc}
  \begin{tabular}{lcccc}
    \toprule
    \ Model  & \multicolumn{3}{c}{Inputs} & Semantic  \\
    \cmidrule{2-4}
     & RGB & Gaze & Gaze & Score ($\uparrow$) \\
     & &(Text)& (Image)& \\
    \midrule
    \multirow{3}{*}{InternVL-2-1B} 
      &  $\checkmark$ &  & &0.1572 \\
      &  $\checkmark$ & $\checkmark$ & & 0.1627 \\
    (one-shot)  & $\checkmark$  &  & $\checkmark$ & 0.1621  \\
    \midrule
    \multirow{3}{*}{InternVL-2-2B} 
     &   $\checkmark$ & & &0.1601  \\
     &   $\checkmark$ & $\checkmark$ & & 0.1682  \\
     (one-shot)&$\checkmark$    & & $\checkmark$ & 0.1694 \\
\midrule
    \multirow{3}{*}{InternVL-2.5-1B} 
     &   $\checkmark$ & & &0.1596  \\
     &   $\checkmark$ & $\checkmark$ & & 0.1692  \\
     (one-shot)&$\checkmark$    & & $\checkmark$ & 0.1679 \\
\midrule
    \multirow{3}{*}{OpenFlamingo-3B} 
     &   $\checkmark$ & & &0.1810  \\
     &   $\checkmark$ & $\checkmark$ & & 0.2000  \\
     (one-shot)&$\checkmark$    & & $\checkmark$ & 0.1991 \\

\midrule
    \multirow{3}{*}{OpenFlamingo-4B} 
     &   $\checkmark$ & & &0.1878  \\
     &   $\checkmark$ & $\checkmark$ & & 0.2025  \\
     (one-shot)&$\checkmark$    & & $\checkmark$ & 0.1998 \\

    \midrule
    \multirow{2}{*}{\textbf{Our method}} 
     &   $\checkmark$ & & &0.6525  \\
     &$\checkmark$    & & $\checkmark$ & \textbf{0.7826} \\
    \bottomrule
  \end{tabular}
  \end{sc}
  \end{scriptsize}
  \end{small}
  \end{center}
  \vspace{-5mm}
\end{table}

\section{Conclusion}
\label{sec:conclusion}

Our work demonstrates that incorporating gaze data significantly enhances VLMs' ability to understand egocentric behavior, improving both predictive capabilities and scene understanding. 
Notably, gaze-regularized models outperform the base model with fine-grained annotations, emphasizing the gaze's role in capturing subtle contextual cues.
Additionally, our study highlights the crucial role of employing a gaze regularizer and a gaze-augmented attention block, providing insights into how to efficiently use and maximize the benefits of gaze information.
Our findings offer a framework that current VLMs can adopt to enhance egocentric behavior understanding. 
To advance research, we will release our code and dataset, to further explore how human gaze can be effectively leveraged to advance VLMs for egocentric behavior understanding and improve human-machine interaction.


\clearpage
\section*{Impact Statement}

We declare that our research does not present any potential ethical issues and is in compliance with the ICML guidelines. The study does not involve human subjects, sensitive data, or methodologies that could result in harmful outcomes or biases. All data this work uses is publicly available, and no privacy or security concerns are implicated.

\bibliography{example_paper}
\bibliographystyle{icml2025}

\onecolumn
\section{Appendix}
\label{appendix} 

In this appendix, 
we present supplementary material related to our study. 
This includes related work, detailed information on the dataset creation process and the design of the prompts used in our experiments. 
We also include results from several ablation studies conducted during our research. 
Finally, we provide essential details about the model training process to assist readers who may wish to reproduce our work from scratch.

\subsection{More Related Work}
\label{sec:more-related-work}

\textbf{Vision Language Models}:
VLMs take in as input images and text together $(image,text)$ and produce a text output. VLMs learn a mapping function $(image,text) \xrightarrow{} text$ where the task varies from Visual Question Answering (VQA) tasks to generating text based on image and text provided. The VLMs output text condition on the image text sequence and several models exists such as BLIP, LLaVa , Flamingo etc. \citet{liu2023visualinstructiontuning}, \citet{li2022blipbootstrappinglanguageimagepretraining},\citet{zhang2023llamaadapterefficientfinetuninglanguage},\citet{alayrac2022flamingo},\citet{chen2023palijointlyscaledmultilinguallanguageimage}.
In this study, we employ the open-source version of the Flamingo model \citet{awadalla2023openflamingo}, built upon the foundations of the original Flamingo described in \citet{alayrac2022flamingo}. The Flamingo is a VLM designed to leverage interleaved text and image data from the internet. It features a pre-trained vision encoder and a trainable perceiver resampler on the vision side. The resampler's learned latent queries are then integrated into the language module. In this module, placeholder tokens for images guide the language model.  These queries, along with the processed text, are input to the cross-attention module which combines the visual features with the language features. The model employs a cross-entropy loss mechanism, aiming to maximize the probability of predicting the correct text token based on preceding image and text tokens. We also utilize InternVL \citep{chen2024internvl} for our experiments since it's open source and have tremendous performance over basic VLM tasks.

\textbf{Action prediction and forecasting}: 
Action prediction and activity forecasting both aim to anticipate future human actions, though they differ in scope. Action prediction focuses on immediate next actions given an input sequence, whereas activity forecasting predicts broader human activities over a longer time frame \citep{gao2017red}. Despite their differences, both tasks share the goal of modeling human behavior to make informed predictions.
These tasks have been widely explored, particularly in the Ego4D benchmark \citep{grauman2022ego4d}. Prior methods have leveraged Long Short-Term Memory (LSTM) networks and modality attention mechanisms \citep{furnari2019expect} to predict future actions. Other studies have adopted bottom-up and top-down approaches to infer latent goals and model temporal dynamics for action forecasting \citep{zhao2024antgptlargelanguagemodels}. \citet{mascaro2024intentionconditioned} decomposed the prediction task into low-level action forecasting conditioned on high-level intentions, improving overall performance.
Hierarchical approaches have also been explored, such as \citet{ashutosh2023hiervllearninghierarchicalvideolanguage}, who used contrastive learning to model both short-term actions ("what the person is doing right now") and long-term intentions ("what the person wants to do"). Similarly, \citet{cho2024shorttermobjectinteractionanticipation} emphasized the role of object-hand interactions, predicting the next active object before modeling future interactions. Transformers and their variants have further enhanced these predictions by capturing interactions between objects and hands \citep{roy2024interactionregionvisualtransformer}. \citet{lai2024legolearningegocentricaction} aimed to generate action frames in egocentric vision via instructional tuning. In our work, we aim to understand egocentric behavior and provide fine-grained text annotations.

\textbf{Attention and gaze-augmented models}:
Attention-based models are widely used to identify important features and improve performance across various domains, including autonomous driving \citep{8082802}, action prediction, and human-computer interaction \citep{9340893, 8793804, 9473584}. These models enhance interpretability by directing focus to the most relevant regions of an image or sequence.
For instance, class activation maps have been employed to leverage pooling layers in deep networks, generating class saliency maps \citep{sudhakaran2018attention} that highlight key objects associated with future actions. Guided-attention mechanisms have further been used to model next-active object interactions, improving action anticipation \citep{thakur2023enhancingactiveobjectbasedegocentric}. The Spatiotemporal Attention Module (STAM) \citep{9022139} incorporates eye gaze as supervision, training a network to predict attention maps for activity recognition. Other studies have similarly used gaze to guide attention maps in activity recognition tasks \citep{Min, awale2022human}.
Beyond artificial systems, human perception and action have long been studied in relation to gaze. Eye movements provide task-specific visual cues, enabling more efficient action execution \citep{Hayhoe2003-gy}. In our work, we explore how eye gaze data can be integrated into Vision-Language Models (VLMs) to improve egocentric behavior understanding, conditioning text annotations on gaze data to enhance future activity prediction.
In our project, we study various ways to utilize eye gaze data in a VLM setting, building a gaze-regularized attention mechanism. Our model conditions predictions on eye gaze data as an input signal to output fine-grained text annotations of events happening in the near future, as well as providing detailed annotations of the current activity.

\subsection{Dataset}
The Ego4D and Epic-Kitchens datasets consist of egocentric videos capturing camera-wearers performing daily activities in controlled settings \citep{grauman2022ego4d,Damen2022RESCALING}.
Other relevant datasets include the EGTEA+ Gaze dataset, which offers 28 hours of content focused on cooking and kitchen activities \citep{li2020eyebeholdergazeactions}, 
and the more recent Visual Data Experience (VDE) dataset, 
which contains around 240 hours of clips documenting day-to-day activities coupled with gaze and head tracking \citep{greene2024visualexperiencedataset200}.

Some datasets are supplemented with coarse-grained annotations or action labels in the form of (noun,verb) but the Ego4D dataset with the gaze modality only contains supplemental gaze data (and some narrations and summaries).
For our study, we collected visual eye gaze data from the Ego4D dataset and obtained more descriptive fine-grained annotations. 
In the future, we plan to incorporate the more recent VDE dataset into our study as well. In addition, \cite{huang2025egoexolearndatasetbridgingasynchronous} provide a dataset which contains egocentric activities as well as an exocentric view. Since humans learn by observing and imitating an expert who demonstrates the actions involved in the activity, utilising this "observe first imitate next" paradigm can probably enhance the downstream tasks as well. 

In this section, 
we provide details about the dataset creation process used for model training and testing, the prompt design, and alternative sources of annotations in the form of VLMs.

\subsubsection{Dataset construction and prompt design}
\label{sec:app_ds_construction}

The Ego4D dataset comprises egocentric video clips along with supplementary data such as audio, text annotations, eye gaze data, and additional metadata \citep{grauman2022ego4d}. 
For our project, 
we focused on the subset of video clips that include eye gaze data, 
containing approximately 33.3 hours of egocentric videos recorded from 80 participants.
The eye gaze data is provided in numerical form, containing canonical timestamps and the pixel coordinates of gaze points. 
We transform gaze points to images to represent important visual regions, 
aligning with how humans perceive spatial information \citep{LAENG2014263}. 
Due to the minute differences between consecutive images in the original videos, 
we perform downsampling to one image per second to reduce computational requirements while remaining effective.
Since our focus is fine-grained egocentric behavior understanding, 
we augment existing annotations with detailed textual descriptions to enhance human-machine interaction. 
We leverage GPT-4V to generate annotations for video frames by processing a sequence of images and prompting it to describe each frame. 
This method ensures contextual coherence across frames. 
After obtaining initial descriptions, we manually evaluate their quality and provide feedback to refine the output. 
This feedback is used to modify the prompt, which is then fed back into GPT-4V with the image frames. 
We conduct prompt selection and refinement using multiple sequences from different video clips to ensure generalizability. 
This iterative process continues until we establish an optimal prompt template that consistently yields accurate and contextually appropriate annotations, confirmed through human evaluation.

\begin{figure*}[t]
  \centering
    \includegraphics[width=1\textwidth]{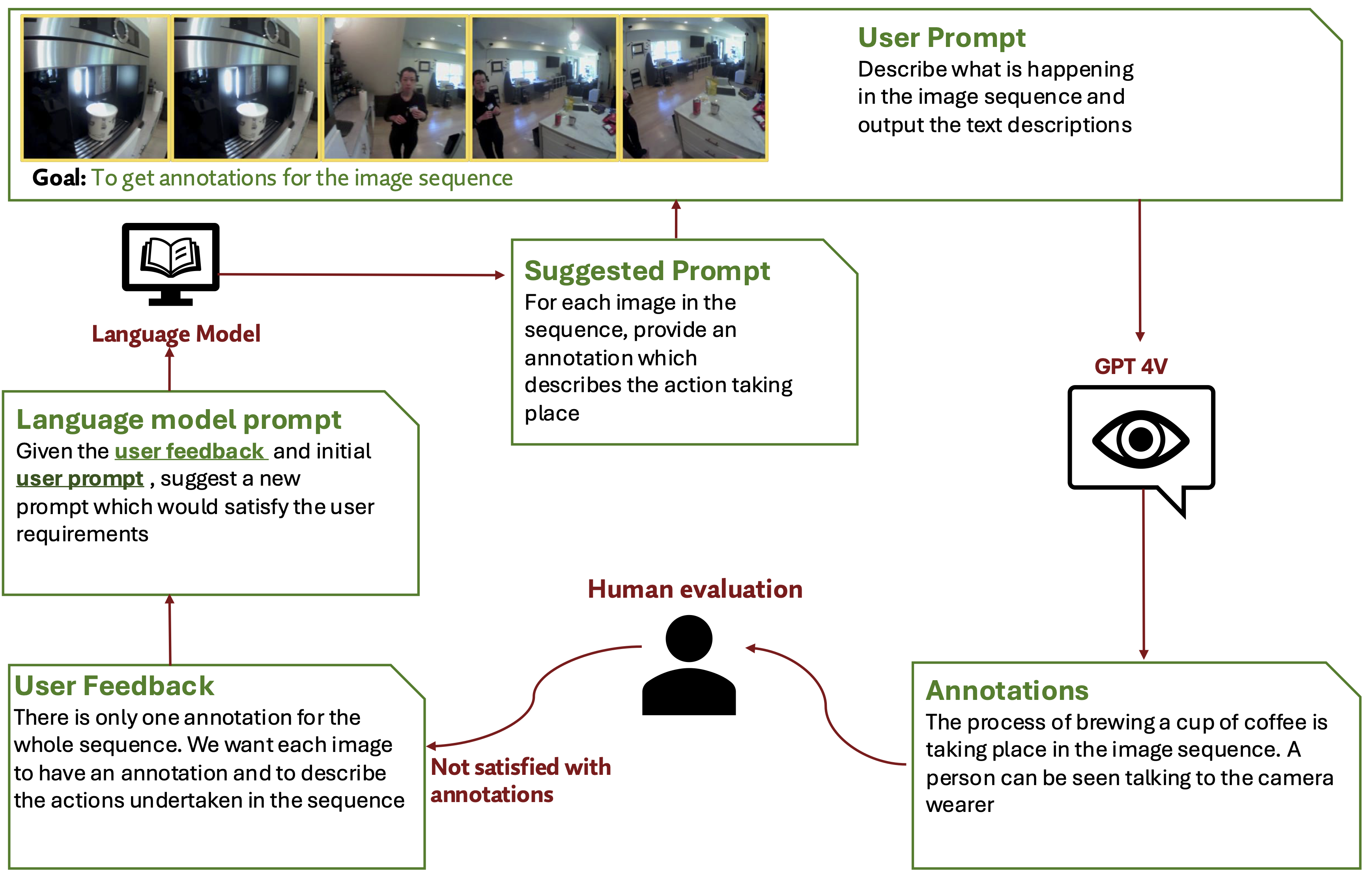}
        \vspace{-8mm}
  \caption{
  \textbf{GPT-4V Iterative Prompting Workflow.} The process begins with a sequence of images and an initial prompt, 
  which are input into GPT-4V to generate annotations for each image. 
  The user then evaluates these annotations and provides feedback, which is incorporated into the prompt using a language model. 
  This modified prompt is used to refine the annotations in a continuous cycle. 
  The objective is to improve the quality and relevance of the output with each iteration until the user is satisfied. 
  For GPT-4V, a set of 10 images is provided at once, ensuring that the annotations maintain contextual coherence.}
  \label{fig:prompt-design1}
\end{figure*}

\begin{figure*}[t]
  \centering
    \includegraphics[width=1\textwidth]{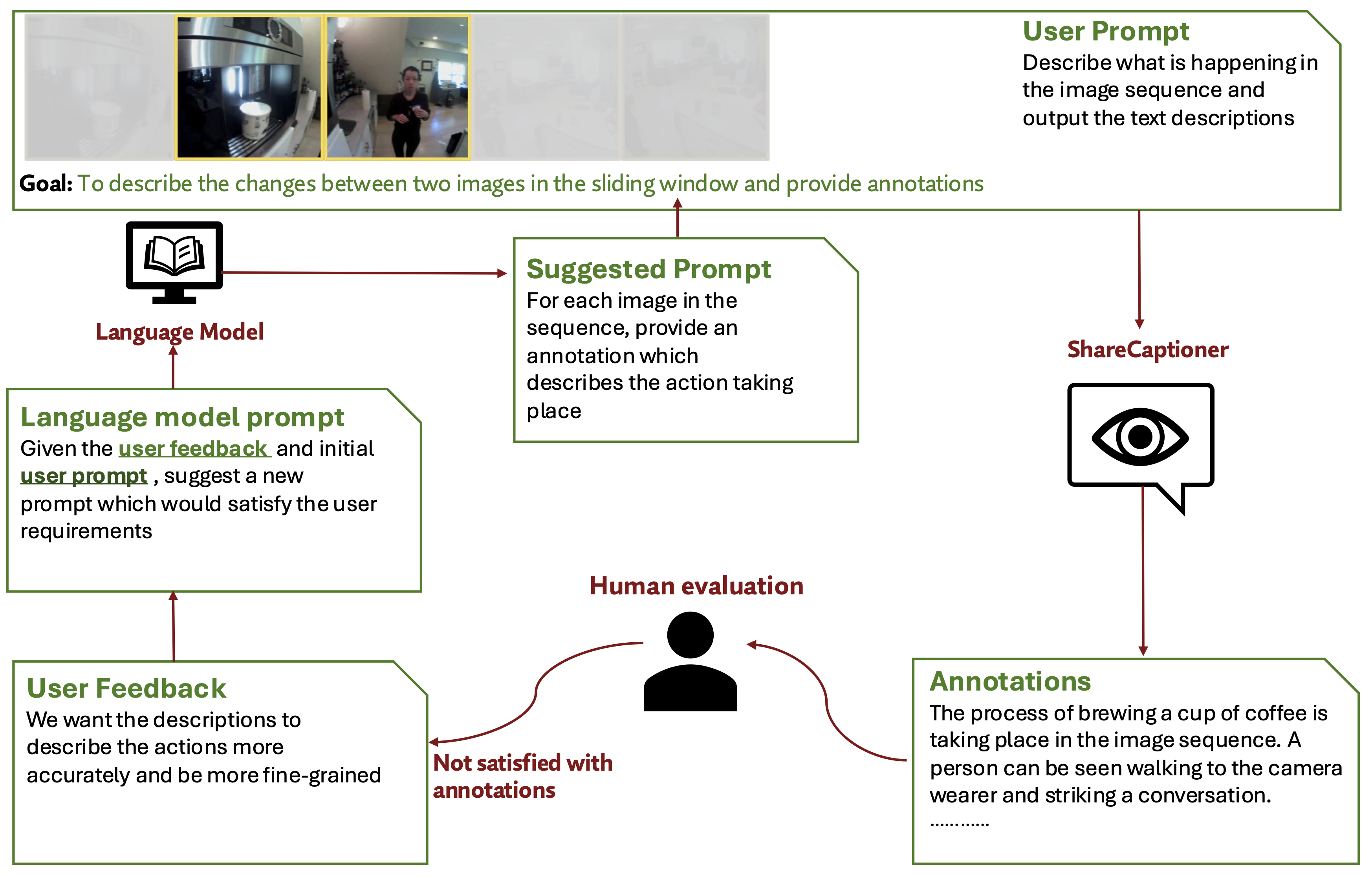}
        \vspace{-8mm}
  \caption{\textbf{ShareCaptioner Iterative Prompting Workflow.} 
  The process begins with a sequence of images accompanied by a starter prompt, 
  which is processed by ShareCaptioner to generate initial annotations. 
  The user then reviews these annotations and provides feedback. 
  This feedback is integrated into the prompt using a language model. 
  The updated prompt is subsequently used to enhance the annotations through an iterative cycle. 
  This cycle continues until the output meets the user’s satisfaction, focusing on improving both quality and relevance. 
  Additionally, a sliding window mechanism is employed to traverse the image sequence, capturing and describing changes as they occur to ensure that all actions are accurately recorded.}
  \label{fig:prompt-design2}
\end{figure*}

Specifically, we start by passing a small set of images to GPT-4V with a basic prompt like: ``Describe what is happening in the image sequence and output the text descriptions.'' 
The initial output was manually evaluated, and feedback was provided. 
This feedback, along with the original prompt, was passed to ChatGPT to refine the prompt for generating accurate and meaningful text annotations. 
The manual evaluation process ensured that the generated annotations met the following criteria:
\begin{enumerate}
    \item A clear description of the objects being manipulated or focused on in the scene.
    \item A detailed account of the actions being performed by both the camera wearer and other individuals present in the images.
    \item Information about any trajectories or movements that take place within the scene.
    \item Clear, fine-grained annotations that fulfill these criteria in a way that is easily understandable by both humans and machines, such as robots that may use these instructions for task execution.
\end{enumerate}
After several iterations of refining the prompt and evaluating the results, 
we identified a suitable template for generating high-quality text annotations. 
This final template was then used to annotate the image sequences using GPT-4V. 
More details on the prompt design process can be found in Figure~\ref{fig:prompt-design1} and Figure~\ref{fig:prompt-design2}.

\subsubsection{Text annotations using an alternate VLM}
In the previous section, 
we obtained text annotations using GPT-4V. 
In addition, we conduct further experiments using an alternate VLM to generate text annotations for the same video clips. 

Specifically, we employed ShareCaptioner \citep{chen2024sharegpt4video} to annotate the entire dataset. 
Under similar training conditions, we compared the empirical performance of the models using annotations from GPT-4V and ShareCaptioner.
The purpose of this comparison was to assess whether there is a significant performance variation -- either a drop or improvement -- when text annotations are sourced from a different VLM. 
Additionally, we aimed to confirm whether the trend of superior performance in gaze-incorporated models persists regardless of the annotation source. 
Finally, we sought to investigate if the granularity of the annotations impacts the model's performance.

\begin{figure*}[t]
  \centering
    \includegraphics[width=1\textwidth]{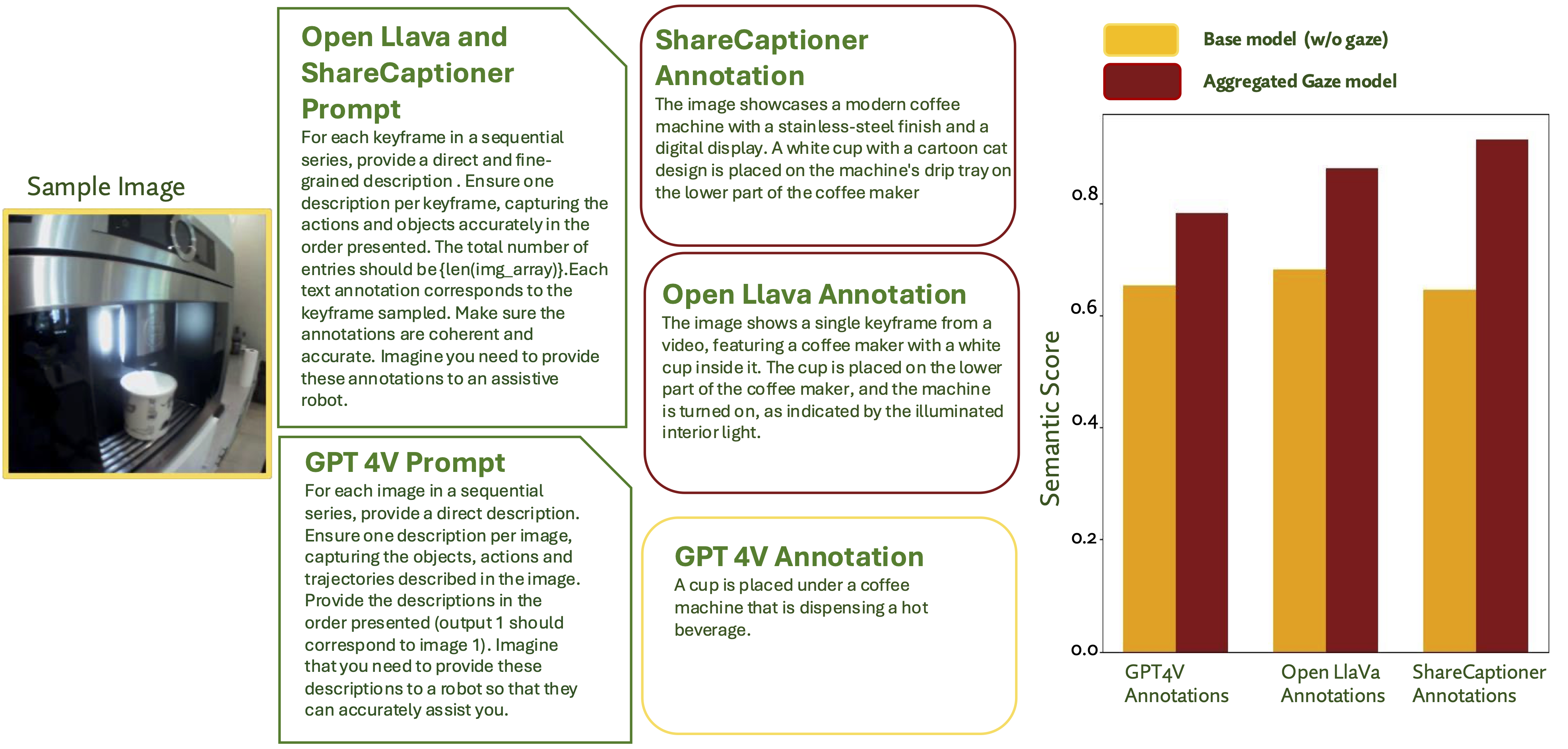}
        \vspace{-8mm}
  \caption{\textbf{Left}: Examples of annotations generated for a sample image by three different annotators: ShareCaptioner, Open Llava, and GPT-4V. The annotations from ShareCaptioner and Open Llava are more fine-grained compared to those produced by GPT-4V in our experimental setup. \textbf{Right}: A graph comparing the performance of the base model and gaze-regularized models, illustrating the impact of using text annotations of varying granularity.}
  \label{fig:prompt-example}
\end{figure*}

As part of this investigation, we also experimented with an open-source implementation of LLaVA to generate finer-grained annotations \cite{chen2024open}. 
However, upon manual evaluation, we found no significant difference between the annotations produced by ShareCaptioner and those generated by LLaVA. In order to have a dedicated comparison between fine-grained annotations and finer-grained annotations on the base model and the gaze-regularized model, we decided to proceed with GPT-4V annotations and ShareCaptioner annotations. However, a comparison of the semantic scores obtained by the base model and aggregated gaze-regularized models for all three annotation models can be found in Figure~\ref{fig:prompt-example}. 
We observed that the gaze-regularized models exhibited a 10-12 percent improvement in the future activity prediction task  when utilizing the finer-grained annotations, 
indicating that more detailed descriptions can further enhance model accuracy in predicting future actions.

Examples of the annotations obtained for a sample image, 
as well as the semantic scores for the base models and the gaze-regularized models, can be seen in Figure~\ref{fig:prompt-example}. 
These examples highlight the difference in performance under varying quality of the fine-grained text annotations.

\textbf{Note:} It is important to emphasize that our goal is not to compare the performance of ShareCaptioner and GPT-4V, 
but rather to evaluate the quality of the annotations obtained from each system and the effect of the quality on the gains of employing gaze in the prediction. 
The differences in annotation granularity may be influenced by cost and time constraints, 
as the setups used to generate these annotations varied. 
Therefore, a direct comparison of the VLMs themselves would be inappropriate.

\subsubsection{Other Information}
We also provide additional details regarding the dataset used in our work. To reiterate, the purpose of using a third-party captioning process was to enhance the granularity of the usual action describing annotations i.e make it more fine-grained. We had approximately 33 hours of data and since the images are sampled at every 1 second, we had a little over 108000 images.The images were divided into 'chunks' each of about size 10 such that when they are provided to GPT-4V, GPT-4V is aware of the full context. Due to connection issues, sometimes we would get an error and so for such cases, we would discard the chunk and so we lost around 5000 images. The annotations for GPT-4V vary from 15-20 words per annotation. The words 'individual' and 'camera-wearer' were the most common and some of the failure cases could be resulting due to the excess amount of the use of the above mentioned words. Two human evaluators were responsible for iteratively refining the GPT-4V prompts and verifying the quality of generated annotations. They evaluated five sets of ten images from ten chosen clips (i.e., 50 images per set, 500 images total) to identify where the annotations were inaccurate or incomplete. Priority was given to identify and flag instances where the annotations are completely wrong and irrelevant. This iterative prompt-refinement and evaluation process took approximately 1.5 weeks in total, including setup time for GPT-4V/ShareCaptioner (how to retrieve the annotations, code etc) and prompt-design experimentation. 

\subsection{Model Overview and Components}
\label{sec:overview-appendix}

In our study, we employ the open-source version of the Flamingo model \citep{awadalla2023openflamingo}, built upon the foundational work of the original Flamingo developed by \citet{alayrac2022flamingo}. The Flamingo is a VLM designed to leverage interleaved text and image data. It features a pre-trained vision encoder to extract input features, a trainable perceiver resampler for creating fixed-size representations of the input features, and trainable cross-attention layers on the language side. The latent features learned by the perceiver resampler are integrated into the language module.
These features, along with the processed text, are input to the cross-attention layers in the language module which combines the visual features with the language features. 
The model utilizes a cross-entropy loss mechanism, aiming to maximize the probability of predicting the correct text token based on preceding image and text.

In addition to traditional input modalities, such as RGB images, recent advancements have highlighted the potential of integrating various modalities beyond vision, including gaze, gait, and tactile sensors \citep{Boshoff2024GaitDP,yang2024bindingtoucheverythinglearning}. Building on this premise, our approach incorporates eye gaze as an additional signal. The use of a transformer incorporated models for egocentric videos was initially proposed by \cite{lai2024eyetransformergloballocalcorrelation} for egocentric gaze estimation. Taking inspiration from the transformer incorporated architecture , we make changes to create a VLM. Specifically, we introduce a gaze-regularized attention mechanism to integrate eye gaze into VLMs.
The base model without eye gaze data serves as a benchmark and relies exclusively on RGB images for input. In contrast, the gaze-regularized model integrates eye gaze data alongside the RGB images. The key distinction of the gaze-regularized attention models is their ability to condition the output using both RGB images and eye gaze data.

In the following section, we provide more details about some of the components in the model architecture, as well as provide information about the occlusion check which is used for gaze point correction during aggregation of gaze points.

\subsubsection{Vision Transformer Encoder}
\label{sec:vit-details}
\begin{figure*}[t]
  \centering
    \includegraphics[width=0.9\textwidth]{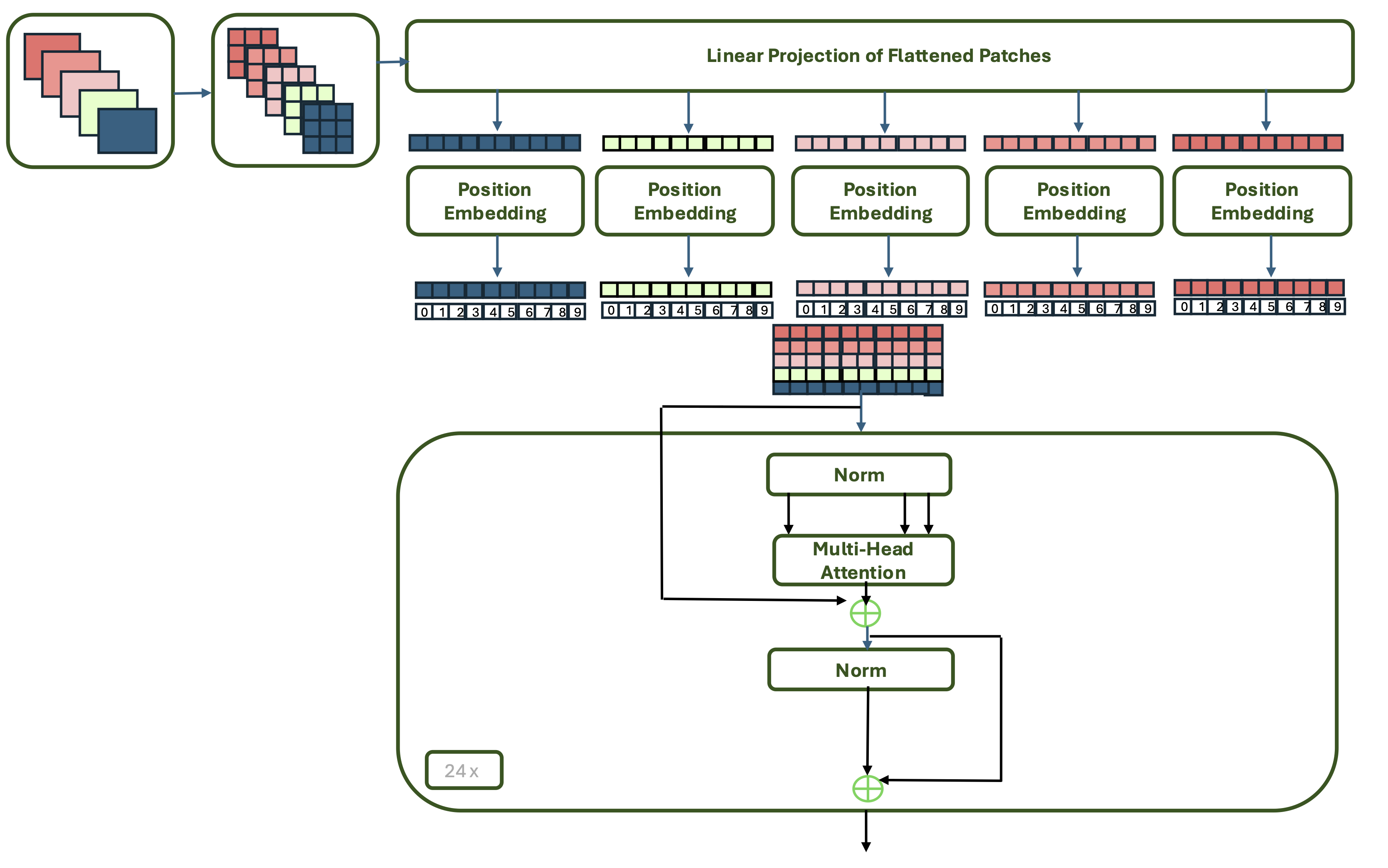}
    \vspace{-5mm}
  \caption{An example of how images are processed by the Vision Transformer in our model. In this case, each image is tokenized into 9 patches. Positional embeddings are then added to the patches, which are subsequently passed through an attention module. In this model, the final MLP head is omitted, as we are focused solely on extracting image features for further analysis.}
  \label{fig:Vit-check}
\end{figure*}

In both the base and gaze-regularized models, we utilize a pre-trained Vision Transformer (ViT) as the image encoder. Images are processed as sequences, where each image is tokenized into patches that are flattened and then transformed into embeddings. To maintain spatial relationships between the patches, these embeddings are supplemented with positional embeddings, ensuring coherence across the entire image.
An illustration of the ViT architecture is provided in Figure ~\ref{fig:Vit-check}. In the figure, as an example, we show how an image is divided into nine tokens, each of which is assigned a positional embedding. These tokenized patches are then passed through the pre-trained transformer encoder, resulting in an embedding that represents the original sequence of image patches. The pre-trained vision encoder employed in our study is ViT-L-14, which results in the formation of 256 tokens for each image.
In the example shown in Figure ~\ref{fig:Vit-check}, the observation length ($\tauo$) is set to 5 seconds, meaning that the image sequences span a 5-second interval. The primary distinction between the original design and the approach used in our study is the omission of the final MLP head, as our focus is exclusively on extracting image features.

\subsubsection{Gaze-guided attention block}
\label{sec:gaze-attn-appendix}
In our study, we propose the inclusion of a gaze-regularized attention mechanism, which features a gaze-guided attention block that operates in conjunction with the gaze regularizer.
In the gaze-guided attention block, gaze-based queries are used for the attention module. These queries contain information about both the scene and the regions occupied by gaze, leading us to refer to this block as the gaze-guided attention block.
The visual features obtained from the RGB image frames, after passing through the ViT, are used as keys and values within the gaze-guided attention block. Similar to the gaze-based features , the features obtained from the RGB image frames pass through a linear layer to form the keys and values for the attention mechanism. The final output, consists of gaze-enhanced features that are informed by the gaze-based queries.

\begin{figure*}[t]
  \centering
    \includegraphics[width=0.9\textwidth]{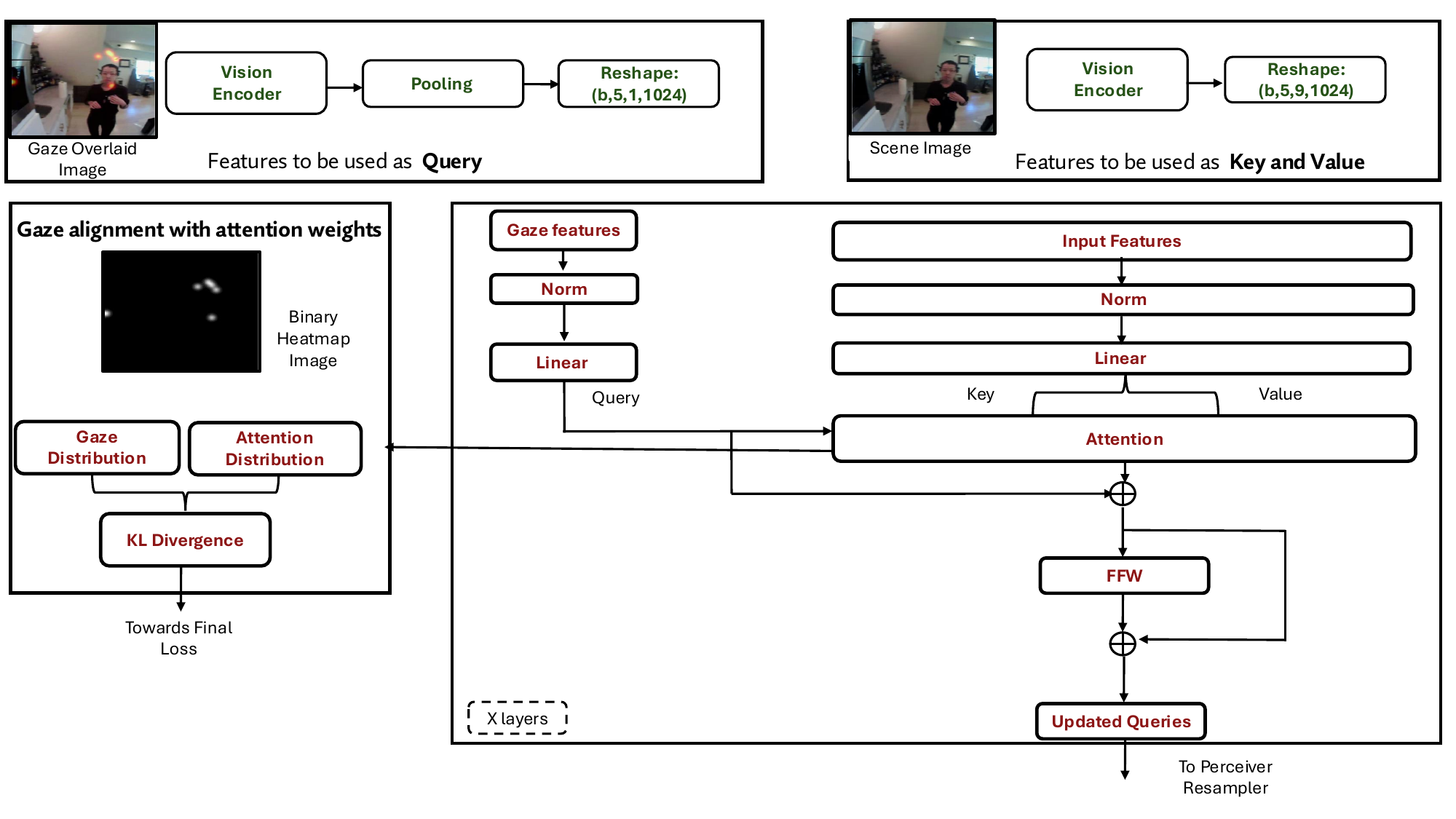}
    \vspace{-4mm}
  \caption{Closer look at the gaze-guided attention block in the pipeline. Gaze overlaid images are passed as queries, whereas the corresponding RGB images are passed as key and value. The binary heatmap image is used to obtain the gaze distribution, which is used in the regulariser by comparing it with the attention distribution obtained from the gaze-guided attention block. The regulariser attempts to align the attention distribution more towards the gaze distribution.}
  \label{fig:attention-block}
\end{figure*}

Additionally, the attention distribution is calculated, reflecting how attention is allocated across all tokenized patches. Initially, this attention distribution may not align with the gaze distribution. Therefore, through regularization, we aim to progressively align the attention distribution with the gaze distribution over time. A closer look at the gaze-guided attention block can be found in Figure~\ref{fig:attention-block}.
    
\begin{figure*}[t]
  \centering
    \includegraphics[width=0.9\textwidth]{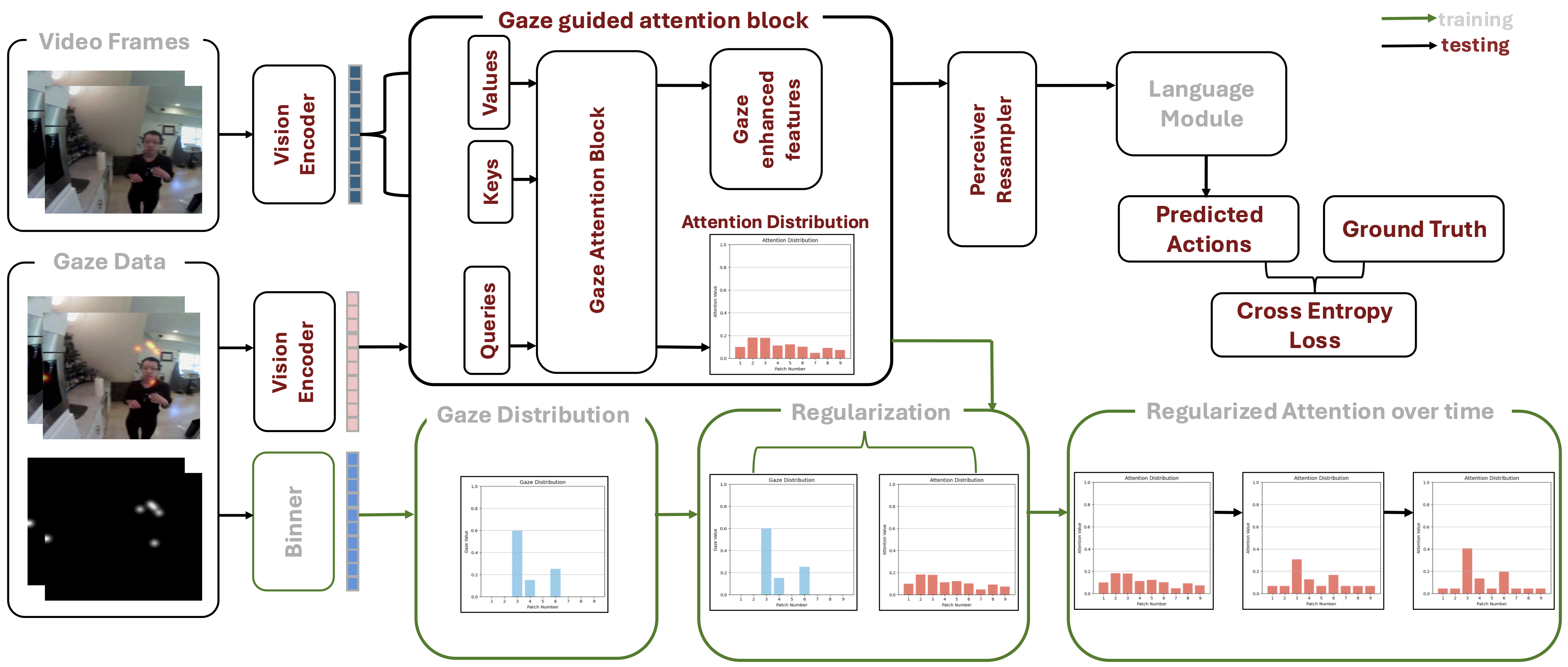}
    \vspace{-4mm}
  \caption{The architecture for the gaze-overlaid supported model which takes in RGB scene image and gaze-overlaid images as input and outputs fine-grained text annotations. The gaze-overlaid supported model derives gaze-based features from the gaze-overlaid images. }
  \label{fig:gaze-arch-block}
\end{figure*}

\subsubsection{Gaze aggregation with occlusion correction}
\label{sec:occlusion-check-correction}
The input image sequence can exhibit dynamic changes due to environmental movement or the movement of the camera wearer. 
The method for aggregating gaze points is suitable only when the frames within the $\delta$ interval for which aggregation is done, shows moderate movement. 
However, preventing movement in a dynamic environment is challenging. 
If we aggregate gaze points in the $[t-\delta, t]$ interval to construct the heatmap $\heat_t$ and there is major occlusion between the earlier frames $[t-\delta, t)$ and the final frame at timestamp $t$, it becomes impractical to include gaze points from the occluded frames. In such cases, collecting gaze points from occluded frames may lead to inaccurate or misleading representations in the heatmap.
To alleviate this issue, 
we perform an occlusion check between each frame in the $[t-\delta, t)$ interval and the final frame at time $t$ to ensure appropriate gaze aggregation. 
In the case of significant occlusion or a drastic change in the scene, 
gaze points corresponding to the earlier frames should not be collected for the aggregation and subsequent formation of the heatmap $\heat_t$.

Using a method similar to the consistency check with optical flow presented by \citet{Hur_2017_ICCV}, 
we explicitly exclude gaze points that are occluded in the current frame. 
If a pixel is correctly translated and there is no major occlusion, then the difference between the forward optical flow displacement of this pixel $(x,y)$ and the 
displacement of the translated pixel $(\hat{x}, \hat{y})$ with backward optical flow should be close to zero.

For an RGB image $\img_t$ at time $t$, we gather the image frames $\{\img_k\}$ for all $k \in [t-\delta, t)$. 
Let the forward optical flow between images $\{\img_k\}$ and $\{\img_t\}$ in the horizontal direction be denoted by $Fx_{k \rightarrow t}$ and the backward optical flow by $Fx_{t \rightarrow k}$. 
Similarly, $Fy_{k \rightarrow t}$ represents the forward optical flow in the vertical direction, and $Fy_{t \rightarrow k}$ represents the backward optical flow in the vertical direction. 
Let the coordinates of a designated pixel $p_i$ be $(x_i, y_i)$. The new coordinates of the translated pixel in the subsequent frame, using optical flow, are computed as follows:
\begin{equation}
    \begin{aligned}
        \hat{x}_{i} = x_{i} + Fx_{k \rightarrow t}(x_{i})\,, \\
        \hat{y}_{i} = y_{i} + Fy_{k \rightarrow t}(y_{i})\,. \\
    \end{aligned}
\end{equation}

Next, we calculate the distance moved by this designated pixel in the horizontal and vertical directions according to the following equations:
\begin{equation}
    \begin{aligned}
        d_{x_{i}} = \lvert Fx_{k \rightarrow t}(x_{i}) \rvert - \lvert Fx_{t \rightarrow k}(\hat{x}_{i}) \rvert\,, \\
        d_{y_{i}} = \lvert Fy_{k \rightarrow t}(y_{i}) \rvert - \lvert Fy_{t \rightarrow k}(\hat{y}_{i}) \rvert\,.
    \end{aligned}
\end{equation}

If the observed proportion of pixels $\eta_{\text{observed}}$ exceeding the distance discrepancy is more than a predefined threshold $\eta$, \
we conclude that a major occlusion has occurred; otherwise, the occlusion is minor. 
The observed proportion of such pixels $\eta_{\text{observed}}$ is calculated as:
\begin{equation}
    \begin{aligned}
            \eta_{\text{observed}} = \frac{\sum_{i=1}^{h \times w} \mathbb{1}_\mathrm{condition} \left( \sqrt{(d_{x_{i}})^2 + (d_{y_{i}})^2} > \epsilon \right)}{\sum_{i=1}^{h \times w} 1}\,,
    \end{aligned}
\end{equation}
where the denominator represents the total number of pixels in the image.

We disregard the gaze points for frames $\{\img_k\}$ where there is major occlusion with respect to the image frame $\img_t$. 
If the occlusion is minor, the appropriate gaze points $\{g_i\}$ for all $i \in [t - \delta, t]$ are then translated into their new coordinates $\{\hat{g}_i\}$ and collected for the formation of heatmap $\heat_t$ and subsequently, for the gaze-overlaid image $\gaze_t$. 
The transformed gaze points are computed as follows:
\begin{equation}
 \begin{aligned} 
    (\hat{g}_{i})_x =    ({g}_{i})_x + Fx_{k\xrightarrow{}t}( ({g}_{i})_x)\,, \\
    (\hat{g}_{i})_y =    ({g}_{i})_y + Fy_{k\xrightarrow{}t}( ({g}_{i})_y)\,. \\
 \end{aligned}
\end{equation}

\begin{figure*}[h]
  \centering
    \includegraphics[width=0.9\textwidth]{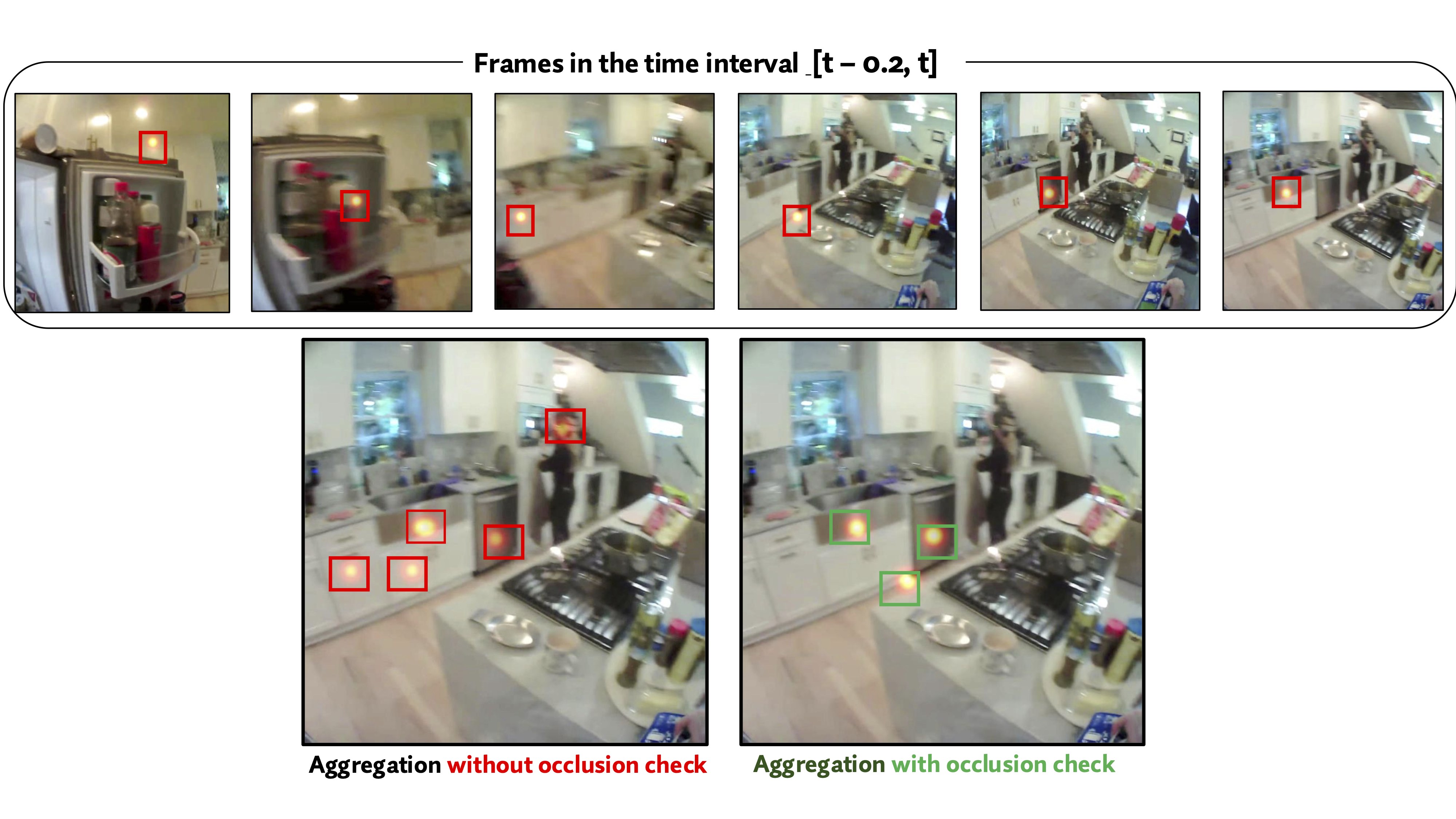}
    \vspace{-4mm}
  \caption{An example of gaze aggregation on the final image frame is shown. \textbf{Bottom left}: Result without occlusion check. \textbf{Bottom right}: Result with occlusion check applied. The fridge is absent in the final frame, indicating significant occlusion. Therefore, the gaze points from the first two frames, which are associated with the fridge, should not be overlaid on the final frame as the object is no longer visible.}
  \label{fig:occlusion-check}
\end{figure*}

As mentioned above, the idea is that if there is a major occlusion, the difference between the distance traversed by a pixel during forward optical flow,and the distance traversed by the translated pixel during backward optical flow will be significantly greater than in cases where the occlusion is minor. 
Optical flow was calculated using the implemntation of the RAFT model developed by \citet{teed2020raftrecurrentallpairsfield}. Human feedback was utilized to distinguish between major and minor occlusions on sample image sequences, which informed the selection of hyperparameters  $\epsilon$ and $\eta$. 
The hyperparameter $\epsilon$ is the threshold distance, which was set to 20, whereas $\eta$ is the threshold proportion of pixels that have exceeded the occlusion limit, set to 0.60.
An example of the result of our occlusion check can be found in Figure~\ref{fig:occlusion-check}. 
It can be observed that if the occlusion check is not present, the aggregation points do not accurately reflect where the person was looking in the final frame on which gaze is overlaid. 
The occlusion check ensures that the correction is effective.

\subsubsection{Alternate ways to obtain Gaze-based queries}
\label{sec:pseudo-gaze-overlaid}
In the previous sections, we used gaze-overlaid images to create gaze-based queries for the gaze attention block. These images are generated by combining the gaze heatmap (constructed using a Gaussian filter and recorded eye gaze points) with scene images. However, it is possible that gaze-overlaid images will not be available in real-time during testing or deployment. Therefore, instead of relying on gaze overlaid images $\{\gaze\}$ to form gaze-based queries, we modify the training setup as follows:
\begin{equation}
 \begin{split}
    \phi_{gaze}(\ell , \{\img\},\{\heat\},\{\gaze_q\}) = 
    \prod_{i=\tau_o+1}^{\tau_o + \tau_a} 
    p(\ell_i \mid \ell_{ < i}, \img_{\leq \tau_o} 
    , {\heat}_{\leq \tau_o},{\gaze_q}_{\leq \tau_o})\,,
 \end{split}
\end{equation}
where $\{\gaze_q\}$ represents gaze-based queries which are obtained without utilizing gaze overlaid images $\{\gaze\}$  and hence gaze-overlaid images are not required during test time.

\textbf{Utilizing gaze-overlaid images as supervision to create gaze-based queries} : We introduce an additional module to generate a heatmap that highlights gaze-relevant regions, which is later combined with the original scene image to create a gaze-overlaid image. The model architecture has two parts:

1) \textbf{Encoder}: Convolutional layers extract high-level features from the input scene image.

2) \textbf{Decoder}: Transposed convolutional layers up-sample the features to produce a gaze heatmap.

\begin{figure*}[h]
  \centering
    \includegraphics[width=1\textwidth]{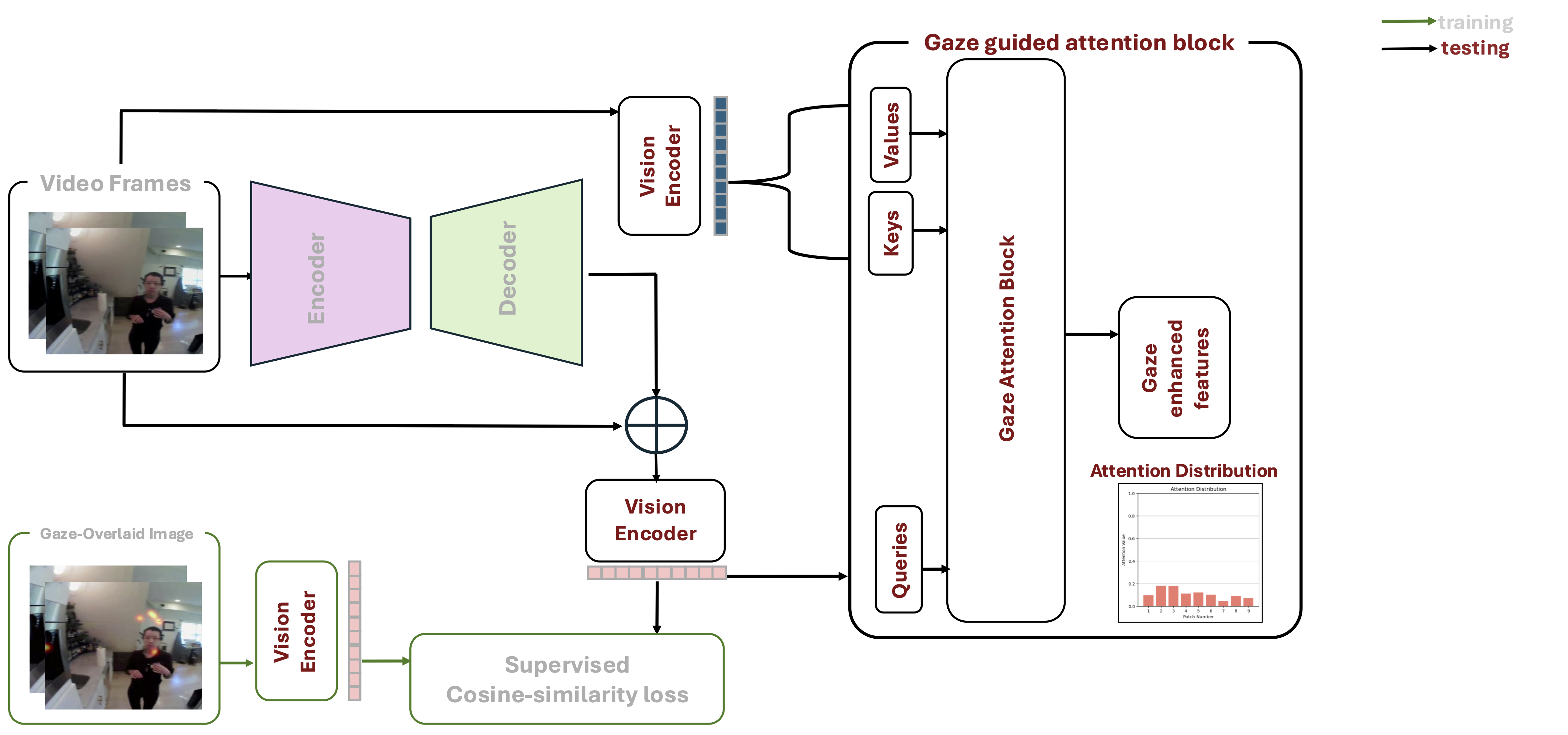}
    \vspace{-4mm}
  \caption{Formation of pseudo-gaze-overlaid images and gaze-based queries. If gaze-overlaid images are not available during test time, an encoder-decoder network dynamically generates a heatmap which is then superimposed on the RGB scene image to form a pseudo-gaze-overlaid image, which is supervised with the ground truth gaze-overlaid image}
  \label{fig:gaze-overlaid-w-supervision}
\end{figure*}

During training, the embeddings from the generated gaze-overlaid image is supervised using cosine similarity loss against embeddings from the ground-truth gaze-overlaid image derived from real eye-gaze data. This ensures the model learns to replicate human gaze patterns accurately. As mentioned above, the module adopts an encoder-decoder architecture designed for extracting features from input scene images and generating a single-channel gaze heatmap. The encoder consists of convolutional layers that progressively downsample the input image while learning high-level features. Specifically, the spatial resolution is halved at each layer, reducing the dimensions to one-quarter of the original size. The decoder then employs transposed convolutional layers to upsample these features back to the original resolution. During training, the generated gaze heatmap is combined with the original scene image to create a gaze-overlaid image.
This approach eliminates the need for recording gaze-overlaid images during deployment. Instead, the model dynamically generates heatmaps and superimposes them on scene-images, enabling real-time, adaptable gaze-overlay generation without external gaze data. Hence, the final objective function becomes:
\begin{equation} 
\mathcal{L}_{\text{total}} 
= \mathcal{L}_{\text{CE}} + \lambda * {D_{KL}} (A \Vert H)\ + \mathcal{L}_{\text{cosine}},
\end{equation}
Figure.~\ref{fig:gaze-overlaid-w-supervision} shows how the heatmaps are generated dynamically whereas the green arrows points towards the components specifically needed during train time. During test time, the gaze-heatmap for supervision is not needed. In this way, we utilize gaze for supervision to predict gazemaps which is then utilized to generate the gaze-based queries, similar to how \cite{Huang_2020} in their study jointly estimated gaze and egocentric actions using a mutual context network. The actions were recognized by a gaze-guided recognition framework while the gaze was predicted using an action-based gaze prediction module.

\subsection{Other Experiments} 
\label{sec:other-exps}
In this section, we provide some details about the ablation studies conducted during the course of our study. 
We first provide an explanation for the metrics used for evaluation and define the evaluation metrics. Then we explore the impact of the size of the gaze points used in the formation of the heatmap on the model performance. In addition, we employ a self-attention block to the base model without gaze to assess the result comparison with the gaze-regularized model. We also utilize gaze in text form to investigate whether our need to use gaze based images is necessary or not, followed by an investigation of the impact of changing observational and prediction horizons on model performance. 

\subsubsection{Evaluation Metrics}

For evaluation, 
we propose using a semantic transformer \citep{reimers2019sentencebertsentenceembeddingsusing} to provide a quantifiable score that compares the generated output with the ground-truth action text, along with the ROUGE-L score.
This scoring system is designed to ensure that the model does not penalize variations of sentences that are semantically similar. For e.g "Football is being played by people" and "People are playing football" are semantically similar, and hence the model should not be penalised too much for such behavior.
Additionally, we aim for the model to penalize sentences where the word order is nonsensical or incomprehensible to humans. 
Alongside the semantic score, we also utilize the ROUGE-L score which is widely employed to assess text similarity. In the main paper, to save table space, ROUGE-L scores are provided with semantic score but tables in the appendix are also supplemented with an additional METEOR score.

\subsubsection{Changes in the size of the overlays of the gaze points}
In our exploration of gaze-regularized models, we investigated two primary approaches: the singular gaze model and the aggregated gaze model. The singular gaze model utilizes only one gaze point to create the binary heatmap and the gaze-overlaid image. To assess the impact of the gaze point size on model performance, we conducted experiments where we increased the size of the overlaid gaze points on the heatmap. Our findings indicated that performance improved with larger gaze point overlays in the singular gaze model.

In contrast, the aggregated gaze model, which combines multiple gaze points and includes an occlusion check, did not exhibit significant performance gains when the size of the overlaid points was increased. Table~\ref{tab:overlay length table} highlights the results of this experiment. The results suggest that the aggregated gaze model's incorporation of multiple gaze points and an occlusion check is more effective than merely enlarging individual gaze points in the singular model. 

\begin{table*}
  \caption{Effect of the size of the gaze point overlays on model accuracy}
  \label{tab:overlay length table}
  \centering
  \begin{sc}
{
  \begin{tabular}{lccccc}
  \toprule 
    \multicolumn{1}{l}{Gaze}& \multicolumn{1}{l}{Semantic} & \multicolumn{1}{l}{Meteor}& \multicolumn{3}{c}{Rouge-L ($\uparrow$)}             \\
    \cmidrule{4-6}
    Model & Score($\uparrow$) & Score($\uparrow$) & Precision & Recall & F-score \\
    \midrule
    Singular   & 0.7316 &0.4501 &0.4822 & 0.5309 & 0.5054    \\
    Singular (larger overlays)  & 0.7683 &0.4922 &0.5123 & 0.5508 & 0.5304    \\
    Aggregated   & \textbf{0.7826} & 0.5033& \textbf{0.5193} & \textbf{0.5644} & \textbf{0.5405} \\
    Aggregated (larger overlays)   & {0.7816}  & \textbf{0.5060} & {0.5128} & {0.5556} & {0.5330} \\

    \bottomrule
  \end{tabular}}
  \end{sc}
\end{table*}

\subsubsection{Gaze-Regularized attention block}

The gaze-regularized attention block 
is a critical addition to the baseline OpenFlamingo model. 
In our ablation studies, we examine the model's performance as the number of gaze-regularized attention blocks varies, 
aiming to identify the optimal number of blocks for further training. 
As shown in Table~\ref{tab:result3}, the performance in the future prediction task improves up to $n=2$, after which it slightly declines with additional blocks. 
Further training confirms that using two gaze-regularized attention blocks yields the best results, as this configuration balances aligning the gaze and attention distributions while preventing over-alignment.


\begin{table}[t]
\caption{Effect of the number of gaze-regularized attention blocks on the performance of the aggregated gaze-regularized models.}
\label{tab:result3}
\vskip 0.15in
\begin{center}
\begin{small}
\begin{sc}
\begin{tabular}{lcc}
\toprule
Attention  & Semantic       & Rouge-L \\
    Blocks             & Score ($\uparrow$)       & F-score ($\uparrow$)    \\
\midrule
1                & 0.7434             & 0.5328          \\
\textbf{2}       & \textbf{0.7826}    & \textbf{0.5405} \\
5                & 0.7765             & 0.5301          \\
\bottomrule
\end{tabular}
\end{sc}
\end{small}
\end{center}
\vskip -0.15in
\end{table}

\subsubsection{Effect of Occlusion Check on Aggregated Gaze-Regularized Models}

In the aggregated gaze model, gaze points are collected within a specified time interval $\delta$ (200 ms). However, in dynamic environments, aggregating gaze points from the interval $[t-\delta, t]$ may introduce inaccuracies due to changes in the scene or camera movement. To mitigate this, we introduced an occlusion check to ensure that only relevant gaze points are aggregated. More details about the occlusion check method can be found in Sec.~\ref{sec:occlusion-check-correction} in the Appendix.

To evaluate the impact of this adjustment, we conducted experiments comparing models with and without the occlusion check in the future prediction task. As shown in Table~\ref{tab:occlusion-check}, the model incorporating the occlusion check slightly outperforms the one without it. The difference in the evaluation metrics can be attributed to the fact that only relevant and accurate gaze points are considered, which reduces noise and prevents the model from being confused by irrelevant data.

\begin{table}[t]
\caption{Comparison of the aggregated gaze-regularized models with and without occlusion check.}
\label{tab:occlusion-check}
\vskip 0.15in
\begin{center}
\begin{small}
\begin{sc}
\begin{tabular}{lcc}
\toprule
Aggregated  & Semantic       & Rouge-L \\
   Gaze Model                   & Score($\uparrow$)       & F-score ($\uparrow$)    \\
\midrule
w/o occlusion-check   & 0.7616             & 0.5286          \\
\textbf{with occlusion-check} & \textbf{0.7826}    & \textbf{0.5405} \\
\bottomrule
\end{tabular}
\end{sc}
\end{small}
\end{center}
\vskip -0.15in
\end{table}

\subsubsection{Inclusion of self-attention block in base model without gaze}

Building on the insights from the previous section, a natural question arises: what happens if we employ a self-attention block in the base model, providing the largest possible overlay—the entire image itself? In this experiment, we integrated a self-attention block into the base model without gaze. The features obtained from the ViT were passed into the self-attention block, where the queries, keys, and values were derived from these image features.

The objective of this modification was to investigate the effects of using a full image overlay as the query input to the attention block in the base model. The resultant features from the self-attention block were then forwarded to the perceiver resampler for further processing. Our results showed that while the performance of the base model improved with the inclusion of the self-attention block, it still remained below that of the gaze-regularized models, as shown in Table~\ref{tab:self-attention table}. This indicates that although leveraging the entire image enhances the base model's capabilities, it does not fully match the performance benefits achieved by incorporating eye gaze signal and using gaze-regularized attention mechanism.

\begin{table*}
  \caption{Comparison of base model with attention block against gaze regularized models}
  \label{tab:self-attention table}
  \centering
  \begin{sc}
{
  \begin{tabular}{lccccc}
  \toprule 
    \multicolumn{1}{l}{}& \multicolumn{1}{l}{Semantic} & \multicolumn{1}{l}{Meteor}& \multicolumn{3}{c}{Rouge-L ($\uparrow$)}             \\
    \cmidrule{4-6}
    Model & Score($\uparrow$) & Score($\uparrow$) & Precision & Recall & F-score \\
    \midrule
   Base &0.6525  & 0.4075  & 0.4335 & 0.4301 & 0.4318    \\
    Base (w self-attention)  & 0.6701 &0.4215 &0.4292 & 0.4508 & 0.4393    \\
    Aggregated gaze   & \textbf{0.7826} & \textbf{0.5033}& \textbf{0.5193} & \textbf{0.5644} & \textbf{0.5405} \\

    \bottomrule
  \end{tabular}}
  \end{sc}
\end{table*}

\subsubsection{Using gaze in text form}

In our studies, we converted gaze data from coordinate text form into heatmaps, which were then overlaid on scene images to create a more visual representation. This transformation allows us to highlight important visual regions and aligns more closely with how humans perceive spatial information. \citep{LAENG2014263}. To conduct a sanity check and assess whether using gaze data in visual form is more suitable than using gaze in text form , we trained a gaze-regularized model that utilizes gaze coordinates as text input. Let $\phi_{gaze,text}$ represent the gaze regularized VLM which aims to model the likelihood of the fine grained text descriptions (represented by $\ell_i$) when eye gaze information in the form of text coordinates $\{(x,y)\}$ is also provided:
\begin{equation}
    \phi_{gaze,text}(\ell , \{\img\},\{(x,y)\})  = \prod_{i=\tau_o+1}^{\tau_o + \tau_a} p(\ell_i \mid \ell_{ < i}, \img_{\leq \tau_o},\{(x,y)\}_{\leq \tau_o})
\end{equation} 

Our results indicated that using gaze data in text form improved performance compared to the base model without gaze. However, it still fell short of the performance achieved by the aggregated gaze-regularized model (as shown in Table~\ref{tab:gaze-text table}). This experiment demonstrates that incorporating gaze information as a signal is essential. However, utilizing gaze in conjunction with the gaze-regularized attention mechanism significantly enhances model performance

\begin{table*}
  \caption{Effect of using gaze information in text form and comparison with other models}
  \label{tab:gaze-text table}
  \centering
  \begin{sc}
{
  \begin{tabular}{lccccc}
  \toprule 
    \multicolumn{1}{l}{}& \multicolumn{1}{l}{Semantic} & \multicolumn{1}{l}{Meteor}& \multicolumn{3}{c}{Rouge-L ($\uparrow$)}             \\
    \cmidrule{4-6}
    Model & Score($\uparrow$) & Score($\uparrow$) & Precision & Recall & F-score \\
    \midrule
   Base &0.6525  & 0.4075  & 0.4335 & 0.4301 & 0.4318    \\
    Aggregated gaze(in text form)  & 0.7021 &0.4428 &0.4642 & 0.4621 & 0.4630    \\
    \textbf{Aggregated gaze}  & \textbf{0.7826} & \textbf{0.5033}& \textbf{0.5193} & \textbf{0.5644} & \textbf{0.5405} \\

    \bottomrule
  \end{tabular}}
  \end{sc}
\end{table*}

\subsubsection{Changing prediction and observation horizons}
\label{sec:horizon_change_obs}
To investigate the effect of sequence length on model performance, 
we conduct experiments by adjusting the prediction horizon ($\taua$) and observation horizon ($\tauo$). 
First, we extend the prediction horizon from 2 to 5 seconds to evaluate both the base and gaze-regularized models (using the aggregated gaze model with occlusion check). 
As shown in Table~\ref{tab:prediction length table}, 
the gaze-regularized model consistently outperforms the base model, 
even with a longer prediction horizon reinforcing the intuition that gaze is important to predict intentions and future actions.
We also reduce the observation horizon to 3 seconds while keeping the prediction horizon fixed at 2 seconds. 
The gaze-regularized model again outperforms the base model (Table~\ref{tab:observation-length table}). 
Interestingly, 
the shorter observation horizon 
shows slightly increased performance compared to the main results reported with a longer observation horizon.
This could indicate a potential improvement of our approach, 
by accounting for the visual working memory humans utilize. 
\begin{table*}[h]
  \caption{On increasing the prediction horizon (from 2 seconds) to predict future actions up to 5 seconds, the gaze-regularized model is able to outperform the base model.}
  \label{tab:prediction length table}
  \centering
  \begin{sc}
{
  \begin{tabular}{lccccc}
  \toprule 
    \multicolumn{1}{l}{}& \multicolumn{1}{l}{Semantic} & \multicolumn{1}{l}{Meteor}& \multicolumn{3}{c}{Rouge-L ($\uparrow$)}             \\
    \cmidrule{4-6}
    Model & Score($\uparrow$) & Score($\uparrow$) & Precision & Recall & F-score \\
    \midrule
    Base  & 0.6297 &0.3972 & 0.4393 & 0.4065 & 0.4220    \\
    \textbf{Aggregated Gaze}   & \textbf{0.7519} &\textbf{0.4891} &\textbf{0.5332} & \textbf{0.5410} & \textbf{0.5386} \\
    \bottomrule
  \end{tabular}}
  \end{sc}
\end{table*}

\begin{table*}[h]
  \caption{On decreasing the observation horizon (from 5 to 3 seconds) to predict future actions up to 2 seconds, 
  the gaze-regularized model is again able to outperform the base model.}
  \label{tab:observation-length table}
  \centering
  \begin{sc}
{
  \begin{tabular}{lccccc}
  \toprule 
    \multicolumn{1}{l}{}& \multicolumn{1}{l}{Semantic} & \multicolumn{1}{l}{Meteor}& \multicolumn{3}{c}{Rouge-L ($\uparrow$)}             \\
    \cmidrule{4-6}
    Model & Score($\uparrow$) & Score($\uparrow$) & Precision & Recall & F-score \\
    \midrule
    Base   & 0.6716 & 0.4284 & 0.4544 & 0.4627 & 0.4581    \\
    \textbf{Aggregated Gaze}   & \textbf{0.7855} &\textbf{0.5132} & \textbf{0.5399} & \textbf{0.5523} & \textbf{0.5457} \\
    \bottomrule
  \end{tabular}}
  \end{sc}
\end{table*}

\subsubsection{Performance when simply gaze overlaid images are provided to base VLM}
To ensure that the performance improvement of our model is not merely due to the additional information provided by the gaze data, we conducted a sanity check. Specifically, we provided gaze-overlaid images as input to the base model instead of standard RGB images. The model was trained under the same conditions as the base model and gaze-regularized models (same number of epochs and same data). The results are provided in Table.~\ref{tab:base-w-gaze-overlaid table}. It is observed that using gaze overlaid images instead of RGB images improves the performance of the base model slightly, but it’s still significantly lower than the performance of the gaze regularized models.

\begin{table*}[h]
  \caption{Comparison for egocentric event prediction when gaze-overlaid images are provided to base VLM.}
  \label{tab:base-w-gaze-overlaid table}
  \centering
  \begin{sc}
{
  \begin{tabular}{lcccc}
  \toprule 
    \multicolumn{1}{l}{}& \multicolumn{1}{l}{Semantic} & \multicolumn{3}{c}{Rouge-L ($\uparrow$)}             \\
    \cmidrule{3-5}
    Model & Score($\uparrow$ & Precision & Recall & F-score \\
    \midrule
    Base w RGB image   & 0.6525 & 0.4335 & 0.4301 & 0.4318 \\
    Base w gaze-overlaid image   & 0.6873 & 0.4332 & 0.4553 & 0.4435 \\
    Aggregated gaze w RGB image   & 0.7826 & 0.5193 & 0.5644 & 0.5405  \\
    \bottomrule
  \end{tabular}}
  \end{sc}
\end{table*}

\subsubsection{Sensitivity to missing data}
During the testing phase, we randomly selected images and corrupted their corresponding gaze-overlaid versions by removing the gaze points entirely. The corruption probability for each gaze-overlaid image was varied to evaluate the model's performance under different levels of corruption. If the corruption probability for an image in the observational sequence is 1, the image contains no gaze overlays (i.e., it is a standard RGB image). The results of this experiment are provided below.
\begin{table*}[h]
  \caption{Comparison for egocentric event prediction when gaze-overlaid images are corrupted}
  \label{tab:gaze-overlaid-corruption}
  \centering
  \begin{sc}
{
  \begin{tabular}{lcccc}
  \toprule 
    \multicolumn{1}{l}{}& \multicolumn{1}{l}{Semantic} & \multicolumn{3}{c}{Rouge-L ($\uparrow$)}             \\
    \cmidrule{3-5}
    Model & Score($\uparrow$ & Precision & Recall & F-score \\
    \midrule
    0   & 0.7826 & 0.5193 & 0.5644 & 0.5405  \\
    0.2   & 0.7801 & 0.5191 & 0.5607 & 0.5390  \\
    0.6   & 0.7757 & 0.5160 & 0.5372 & 0.5264  \\
    1   & 0.7368 & 0.4958 & 0.5113 & 0.5034  \\
    \bottomrule
  \end{tabular}}
  \end{sc}
\end{table*}

\subsubsection{Out-Of-Distribution Data}
While previous experiments used test data from the same dataset as training, we also evaluated our models on egocentric videos from the EGTEA+ Gaze dataset to assess performance on out-of-distribution (OOD) data for future activity prediction. Both the base model and the aggregated gaze model were tested using fine-grained text annotations.

As shown in Table~\ref{tab:ood-data}, the base model performs similarly on OOD data as it does on the original dataset. The aggregated gaze model experiences a slight performance drop but still outperforms the base model by 9 $\%$ and remains the best-performing model overall. This demonstrates the benefit of gaze-overlaid images for improved generalization.

The performance drop in gaze-based models on OOD data is likely due to resizing adjustments needed to match the training format, which may have diminished the impact of gaze information. However, the base model’s consistent performance suggests that in egocentric videos, key events often occur near the center of the frame, making the OOD data less distinct from the original dataset and minimizing the impact of distribution shift.

    
\begin{table}[h]
\caption{Performance of models using out-of-distribution data.}
\label{tab:ood-data}
\vskip 0.15in
\begin{center}
\begin{sc}
\begin{tabular}{lccc}
\toprule
Model                                   & Semantic       & RGB  & Gaze-Overlaid \\
                                        & Score($\uparrow$)  & Input          &Input                     \\
\midrule
Base Model                              & 0.6514             & $\checkmark$ &                     \\
\textbf{Aggregated Gaze}          & \textbf{0.7425}    & $\checkmark$ & $\checkmark$        \\
Aggregated gaze & 0.7005             & $\checkmark$ &                     \\
using RGB queries & & \\
Pseudo-gaze     & 0.7215             & $\checkmark$ &                     \\
supported & & \\
\bottomrule
\end{tabular}
\end{sc}
\end{center}
\vskip -0.15in
\end{table}
\subsubsection{Ablation studies with more gaze points}
Due to constraints in time and resources, we were unable to conduct experiments using 30 gaze points. However, to explore the impact of using a larger number of points (or a longer aggregation duration), we trained an aggregated gaze model utilizing 12 frames instead of 6 (and hence the aggregation time $\delta$ is 400 ms). The results of this experiment are provided in Table.~\ref{tab:gaze-w-more-points}. The negligible difference between the two scenarios can likely be attributed to occlusion checks, which may limit the total number of usable gaze points. In cases with minimal occlusion, the slight decrease in performance can be compared to that observed in an aggregated gaze model with larger overlays, where performance also decreased slightly. This result suggests that utilizing an excessive number of gaze points could potentially confuse the model. However, as this explanation is currently based on intuition rather than extensive analysis, this experiment was not included in the main paper.

\begin{table*}[h]
  \caption{Comparison when more aggregated points are used for the gaze-regularized model}
  \label{tab:gaze-w-more-points}
  \centering
  \begin{sc}
{
  \begin{tabular}{lcccc}
  \toprule 
    \multicolumn{1}{l}{}& \multicolumn{1}{l}{Semantic} & \multicolumn{3}{c}{Rouge-L ($\uparrow$)}             \\
    \cmidrule{3-5}
    Number of points & Score($\uparrow$ & Precision & Recall & F-score \\
    \midrule
    6  & 0.7826 & 0.5193 & 0.5644 & 0.5405 \\
    12  & 0.7808 & 0.5143 & 0.5541 & 0.5334 \\
    \bottomrule
  \end{tabular}}
  \end{sc}
\end{table*}



\subsection{Limitations}
\label{sec:limitations}
While our approach demonstrates the benefits of incorporating gaze into VLMs, it has some limitations. The dataset was created using an iterative prompt refinement process, with template approval by two individuals. Although we manually verified a subset of annotations, the subjective nature of annotations means some inconsistencies may persist. Additionally, due to the lack of multimodal data (e.g., audio), our study focuses solely on visual and gaze information. We emphasize that our goal is not to introduce a new benchmark but to highlight the value of gaze as an additional modality for enhancing egocentric behavior understanding in VLMs. The occlusion check also requires human feedback to fix some of the parameters and this occlusion check component should be further enhanced in the future as its an integral part of the system.

\subsection{Training and evaluation details}
\label{sec:training-details}
Both the base model and the gaze-regularized model were trained using two NVIDIA A800 40GB GPU cards. The base model required approximately 36-38 hours for training, while the gaze-regularized model took around 50 hours. The training utilized a batch size of 32 and a learning rate of $7 \mathrm{e}-5$. The vision encoder is pre-trained by OpenAI. Additionally, the tokenizer for the text annotations used was the OPT (Open Pre-trained Transformer) language model developed by Meta. To accelerate the training process, we employed Fully Sharded Data Parallel (FSDP), a technique that efficiently distributes model parameters and gradients across multiple GPUs, reducing memory usage and improving training speed.
Data loading was managed using the WebDataset loader, and the dataset was converted to .tar files to align with the format required for integration with both the WebDataset loader and FSDP.

In terms of model complexity, the base model has approximately 944 million parameters. The gaze-regularized model with 5 layers contains about 996 million parameters while the gaze-regularized model with 1 layer has around 955 million parameters. Lastly, the model with 2 layers, which is the best performing model, has approximately 966 million parameters. The pseudo-gaze-overlay supported model utilises an addition trainable encoder-decoder network and has 40 million parameters more.

For evaluation, both models were assessed using the semantic transformer (SBERT) developed by \citet{reimers2019sentencebertsentenceembeddingsusing}, along with ROUGE-L scores. The inference time on the test set for the base model was approximately one hour, while the gaze-regularized model had a slightly longer inference time but remained comparable to that of the base model.

Regarding the runtime, we would like to clarify that the evaluation was conducted using RGB images for the base model and RGB+gaze-overlaid images for the aggregated gaze model. On average, both models took approximately 1.7 to 2 seconds to process a sequence. This runtime was calculated based on the total time required to run the test set divided by the total number of samples in the test set.
For the aggregated gaze model, the incorporation of an occlusion check using forward and backward optical flow adds an additional 2-3 seconds per sequence, resulting in a slightly slower runtime compared to the base model. However, this occlusion check improves the quality of predictions, as reflected in the semantic scores.
If the occlusion check (optical flow with consistency) is excluded during test time, the semantic score decreases from 0.7826 to 0.7702, highlighting its contribution to model performance. Additionally, if the occlusion check is removed during training, the semantic score drops further to 0.7616. These results emphasize the importance of consistency checks in improving the model's semantic understanding. In addition, we aim to present the gaze-regularized framework as an adaptable component that can be integrated into Vision Language Models VLM). This framework is designed to work seamlessly with transformer-based architectures that utilize separate vision and language modules, which we think will make it suitable to other VLMs

\end{document}